\documentclass[journal]{IEEEtran}
\ifCLASSINFOpdf
\else
\fi
\usepackage{amsmath}
\usepackage{subfigure}
\usepackage{graphics} 
\usepackage{epsfig} 
\usepackage{mathptmx} 
\usepackage{times} 
\usepackage{amsmath}
\usepackage{amssymb}  
\usepackage{booktabs}
\usepackage{url}
\usepackage{multirow}
\usepackage{bbding}
\usepackage[table,xcdraw]{xcolor} 
\usepackage{bbold}
\usepackage{booktabs}
\usepackage{siunitx}

\begin{document}
%
\title{Achelous++: Power-Oriented Water-Surface Panoptic Perception Framework on Edge Devices based on Vision-Radar Fusion and Pruning of Heterogeneous Modalities}
%
%
%

\author{Runwei Guan$^{\star}$, 
        Haocheng Zhao$^{\star}$,
        Shanliang Yao$^{\star}$, 
        Ka Lok Man, 
        Xiaohui Zhu, 
        Limin Yu,
        Yong Yue, 
        Jeremy Smith, 
        Eng Gee Lim, ~\IEEEmembership{Senior Member,~IEEE},
        Weiping Ding, ~\IEEEmembership{Senior Member,~IEEE},
        and Yutao Yue$^{*}$

\thanks{$^{1}$Runwei Guan, Haocheng Zhao, Shanliang Yao ($^{\star}$Equal Contribution.) and Jeremy Smith are with Faculty of Science and Engineering, University of Liverpool, L69 3BX Liverpool, United Kingdom. 
        {\tt\small \{runwei.guan, haocheng.zhao, shanliang.yao, J.S.Smith\}@liverpool.ac.uk}}
\thanks{$^{2}$Runwei Guan, Haocheng Zhao and Shanliang Yao are with XJTLU-JITRI Academy of Technology, Xi'an Jiaotong-Liverpool University, 215123 Suzhou, China.  
        {\tt\small \{runwei.guan21, haocheng.zhao19, shanliang.yao19\}@student.xjtlu.edu.cn}}
\thanks{$^{3}$Ka Lok Man, Xiaohui Zhu, Limin Yu, Eng Gee Lim and Yong Yue are with School of Advanced Technology, Xi'an Jiaotong-Liverpool University, 215123 Suzhou, China.  
        {\tt\small \{Ka.Man, xiaohui.zhu, limin.yu, enggee.lim, yong.yue\}@xjtlu.edu.cn}}
\thanks{$^{4}$Runwei Guan, Haocheng Zhao, Shanliang Yao, Yutao Yue ($^{*}$corresponding author) are with Institute of Deep Perception Technology, JITRI, 214000 Wuxi, China.  
        {\tt\small \{guanrunwei, zhaohaocheng, yaoshanliang, yueyutao\}@idpt.org}}%
\thanks{$^{5}$Weiping Ding is with School of Information Science and Technology, Nantong University, 226000 Nantong, China.  
        {\tt\small ding.wp@ntu.edu.cn}}%
}

%
%

\markboth{Journal of \LaTeX\ Class Files,~Vol.~14, No.~8, August~2015}%
{Shell \MakeLowercase{\textit{et al.}}: Bare Demo of IEEEtran.cls for IEEE Journals}
%



\maketitle

\begin{abstract}
Urban water-surface robust perception serves as the foundation for intelligent monitoring of aquatic environments and the autonomous navigation and operation of unmanned vessels, especially in the context of waterway safety. It is worth noting that current multi-sensor fusion and multi-task learning models consume substantial power and heavily rely on high-power GPUs for inference. This contributes to increased carbon emissions, a concern that runs counter to the prevailing emphasis on environmental preservation and the pursuit of sustainable, low-carbon urban environments. In light of these concerns, this paper concentrates on low-power, lightweight, multi-task panoptic perception through the fusion of visual and 4D radar data, which is seen as a promising low-cost perception method. We propose a framework named \textit{Achelous++} that facilitates the development and comprehensive evaluation of multi-task water-surface panoptic perception models. \textit{Achelous++} can simultaneously execute five perception tasks with high speed and low power consumption, including object detection, object semantic segmentation, drivable-area segmentation, waterline segmentation, and radar point cloud semantic segmentation. Furthermore, to meet the demand for developers to customize models for real-time inference on low-performance devices, a novel multi-modal pruning strategy known as \textit{Heterogeneous-Aware SynFlow (HA-SynFlow)} is proposed. Besides, \textit{Achelous++} also supports random pruning at initialization with different layer-wise sparsity, such as \textit{Uniform} and \textit{Erd\H{o}s-R\'enyi-Kernel (ERK)}. Overall, our \textit{Achelous++} framework achieves state-of-the-art performance on the WaterScenes benchmark, excelling in both accuracy and power efficiency compared to other single-task and multi-task models. We release and maintain the code at \url{https://github.com/GuanRunwei/Achelous}.

\end{abstract}

\begin{IEEEkeywords}
Water-surface perception, low-power model, multi-task learning, vision-radar fusion, multi-modal pruning
\end{IEEEkeywords}

%
\IEEEpeerreviewmaketitle


\section{Introduction}
\begin{figure*}[ht]
    \includegraphics[width=1\linewidth]{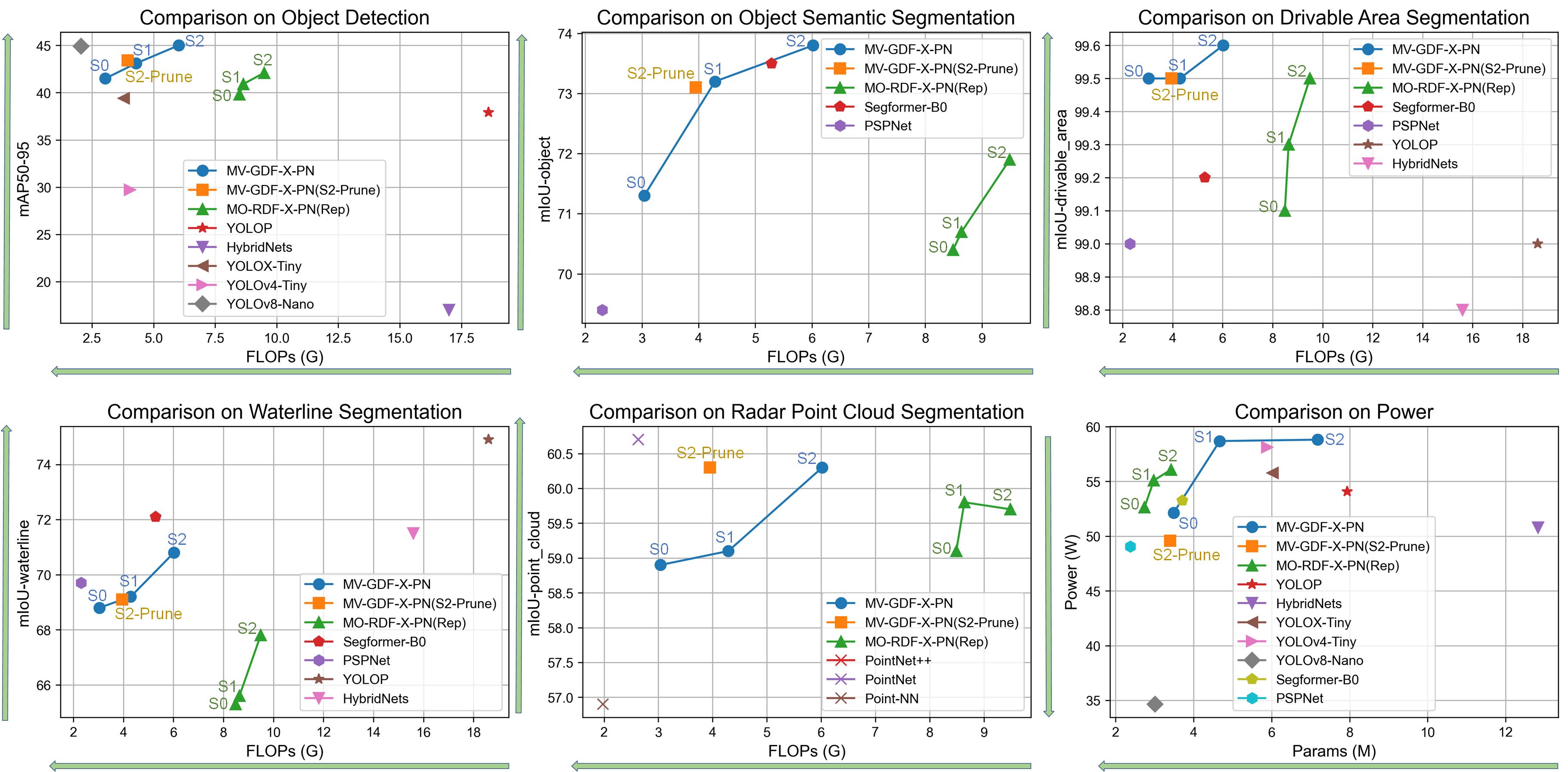}
    \caption{Comparison of selective variants of Achelous++, including MV-GDF-X-PN (S0, S1, S2), MV-GDF-X-PN (S2-Prune) and MO-RDF-X-PN (S0-Rep, S1-Rep, S2-Rep) with various models on different tasks in WaterScenes benchmark.}
    \label{fig:overview_compare}
\end{figure*}
Recently, autonomous driving has witnessed significant progress in robust water-surface environmental perception \cite{cheng2021robust}\cite{cheng2021flow}\cite{guan2023achelous}\cite{guan2023efficient}. This progress is particularly concentrated on multi-sensor fusion \cite{ounoughi2023data} and multi-task learning \cite{vandenhende2021multi}. Multi-sensor fusion primarily focuses on combining data from cameras, lidar, and radar sensors, allowing them to complement each other to achieve robust perception when confronting adverse weather \cite{lxlliu}. Currently, 4D radar-vision fusion is considered a low-cost and promising perception solution \cite{yao2023radar}\cite{yao2023waterscenes}\cite{palffy2022multi}\cite{zheng2022tj4dradset}. In addition, to enable planning-oriented perception \cite{ye2023fusionad}\cite{hu2023planning}, Unmanned Surface Vehicles (USVs) must perform multiple perception tasks concurrently, rather than fixating on a single task. Moreover, this approach not only increases the scope of understanding but also improves individual task performance \cite{qu2023multi}\cite{kendall2018multi}\cite{sener2018multi}. 

However, single-task model parallel systems, as well as multi-task and multi-modal learning, inevitably lead to the generation of more parallel branches and fragmented designs within the model, resulting in increased complexity. This complexity brings about various negative consequences. For instance, the model may not perform real-time inference on low-power edge devices such as NVIDIA Jetson or Orin \cite{cheng2021robust}\cite{guan2023efficient}, necessitating its operation on higher-performance, higher-power devices. Furthermore, the model itself consumes more power than conventional single-task models, contributing to increased carbon emissions. Presently, multi-modal and multi-task perception models predominantly emphasize improving accuracy \cite{guan2023efficient}\cite{ye2023fusionad}\cite{qu2023improving}, seemingly neglecting the environmental impact of deep learning models.

We contend that low-power water-surface perception models have the potential to reduce energy consumption and lower carbon emissions. Conventional high-power perception systems often require more energy supply, while low-power models can reduce power demands, thereby mitigating a city's carbon footprint and advancing low-carbon sustainable urban objectives \cite{tu2023femtodet}. Moreover, low-power models enhance data privacy protection. They process data on edge devices carried by USVs, reducing the need for data transmission and storage on remote servers \cite{peng2022online}. This reduction in data transfer minimizes the risk of data breaches and eliminates the synchronization issues that can result from communication delays, thus preventing autonomous navigation accidents. Furthermore, low-power water-surface perception models typically incur lower maintenance and operational costs, effectively extending their operational uptime.


Upon the aforementioned survey, we contemplate the rationality and significance of a low-cost multi-task environment perception framework based on the fusion of a monocular camera and 4D mmWave radar, especially for edge devices. Here are our considerations, 

\begin{enumerate}
    \item 4D (mmWave) radar is an all-weather robust perception sensor. It is immune to light and adverse weather, which can obtain denser point clouds and more precise elevation measurements. 4D radar can complement visually perceived information well and take over the perception system in time when vision fails \cite{liu2023smurf}.
    \item Multi-task model can wave the consideration of matching and synchronization from single-task standalone models. Multi-task learning can also improve each individual task by optimizing the shared feature space comprehensively \cite{guan2023achelous}, but how to streamline models efficiently without compromising performance is difficult.
    \item Low cost includes the low power consumption, complexity and latency of models, which is significant for perception models. Low-cost models can speed up inference, prolong endurance on edge devices and benefit for low-carbon environment. Nevertheless, most multi-task models exhibit high complexity, and striking a balance between model complexity, speed, and power consumption poses a challenge \cite{tu2023femtodet}.
    \item Currently, most works focus on single-task or single-modal models, instead of frameworks. A framework that is compatible with multiple model structures can help researchers/developers implement their desired models more quickly.
\end{enumerate}

Based on these considerations, we focus on robust urban water-surface perception based on the fusion of vision and 4D radar. Our contributions are as follows,

\begin{enumerate}
    \item A unified 2D multi-task panoptic perception framework called Achelous++. Achelous++ is a low-cost and high-performance framework based on feature-level and task-level fusion of vision and radar point clouds. It achieves state-of-the-art performances (Fig. \ref{fig:overview_compare}) on the real-time panoptic perception of WaterScenes benchmark \cite{yao2023waterscenes}.
    \item Achelous++ is a framework with high degrees of freedom, which supports the replacement and modification of various modules from different stages in neural networks. It simultaneously supports native PyTorch code and encapsulated framework code, such as MMDetection \cite{chen2019mmdetection} or Detectron2 \cite{wu2019detectron2}. Achelous++ also provides a series of evaluation functions to comprehensively evaluate performances of models from perspectives of accuracy, speed, power consumption, etc.
    \item To refinedly extract the features of radar point clouds, we propose radar convolution operator, which is more friendly and fast to the irregularness of radar point clouds on 2D image planes than ordinary convolution.
    \item Aiming at the convenience of developers on slimming the model, we provide a structural pruning library for Achelous++, where a novel heterogeneous modalities fusion-aware salience-based pruning algorithm Heterogeneous Aware SynFlow (HA-SynFlow) is proposed. The algorithm can detect the fusion method, score the salience for each modality, and prune with different modality-wise sparsity.
\end{enumerate}

The remaining content is organized as follows, Section \ref{sec:related} presents the related works; Section \ref{sec:framework} illustrates the detailed motivation and information of Achelous++ design; Section \ref{sec:experiments} provides an abundant and comprehensive experiments and analysis; Section \ref{sec:conclusion} concludes the paper while Section \ref{sec:discussion} discusses the limitation of the paper and expects our future work.

\section{Related Works}
\label{sec:related}

\subsection{Multi-Task Learning for Environment Perception}
It is a tendency to adopt multi-task models to complete the environment perception, which can lower the complexity of the perception system compared with standalone models. Furthermore, multi-task models can improve the performance of individual perception tasks by adopting suitable multi-objective optimization strategies to optimize the model containing shared parameters. YOLOP \cite{wu2022yolop}, HybridNets \cite{vu2022hybridnets} and YOLOPv2 \cite{han2022yolopv2} are three well-known multi-task panoptic perception models, that perform object detection, drivable area segmentation and lane segmentation at the same time. Efficient-VRNet \cite{guan2023efficient} proposes an asymmetric fusion mechanism to improve the performances of detection and segmentation bilaterally. UniAD \cite{hu2023planning} is a great work focusing on planning-oriented 3D perception, which can simultaneously complete detection, tracking, mapping, motion prediction and occupancy prediction. 

In addition to multi-task model architectures, multi-task optimization strategies matter \cite{lin2023libmtl}. Uncertainty weighting \cite{sener2018multi} is a weighting-based strategy based on homoscedastic uncertainty for multi-task loss weighting. MGDA \cite{kendall2018multi} explicitly defines multi-task learning as multi-objective optimization, where the overall goal is to find Pareto optimal solutions. Aligned-MTL \cite{senushkin2023independent} is a novel and state-of-the-art multi-task optimization method that eliminates instabilities during training by aligning the orthogonal components of a linear gradient system.

\subsection{2D Driving Perception Based on Vision-Radar Fusion}
2D perception is a low-cost approach compared with 3D perception, which mainly includes object detection, drivable-area segmentation, lane segmentation, panoptic segmentation and instance segmentation. CRFNet \cite{nobis2019deep}, using VGGNet as backbones, is a classical object detection network based on the feature-level fusion of image and radar, \cite{stacker2022fusion}\cite{kumawat2022radar}\cite{li2020feature}\cite{kowol2020yodar} are four works following the paradigm of CRFNet through the addition fusion of image and radar features. RVNet \cite{john2019rvnet} and RISFNet \cite{cheng2021robust} are two fusion networks for object detection by concatenation of image and radar features. Efficient-VRNet \cite{guan2023efficient} is a multi-task perception network based on a novel asymmetric fusion between image and radar modalities, which can simultaneously perform object detection, semantic segmentation and drivable-area segmentation. Based on the above survey, we find these researches mainly focused on object detection only and the models are relatively complex, reflecting on model size and FLOPs, which are not friendly to edge-end deployment. Furthermore, only CRFNet and Efficient-VRNet are open-source and it is tough for researchers to evaluate performances of other models.


\subsection{Structural Pruning and Pruning at Initialization}

Researchers have proposed various model compression algorithm for deployment, including pruning, quantization, distillation, lightweight kernel, and reparameterization. Among these, pruning stands out as one of the most fundamental and efficient methods \cite{gao2023significant}. As early as 90s in last century, Lecun et al. introduced algorithms like OBD \cite{lecun1989optimal} and OBS \cite{hassibi1993optimal}, employing second-order gradient-based techniques for unstructural pruning. However, unstructural sparsity relies on sparse acceleration in specific hardware, making it challenging to accelerate model inference. Structural pruning, on the other hand, could significantly accelerate neural network inference by reducing weights along specific dimensions. While the Lottery Ticket Hypothesis (LTH) \cite{frankle2018lottery} has had a significant impact in the pruning domain, its effects on structural pruning are less pronounced and might even extend training times \cite{liu2018rethinking}. The concept of Pruning at Initialization (PaI) has shown great promise in recent years, achieving notable sparsity and precision levels with nearly the same computational costs as regular training \cite{liu2022unreasonable}. Therefore, this paper utilizes an PaI structural pruning strategy to offer a lightweight version of the Achelous++ models. This approach effectively accelerates model inference speed while maintaining a satisfied performance.

\begin{figure}[ht]
\begin{center}
    \includegraphics[scale=0.2]{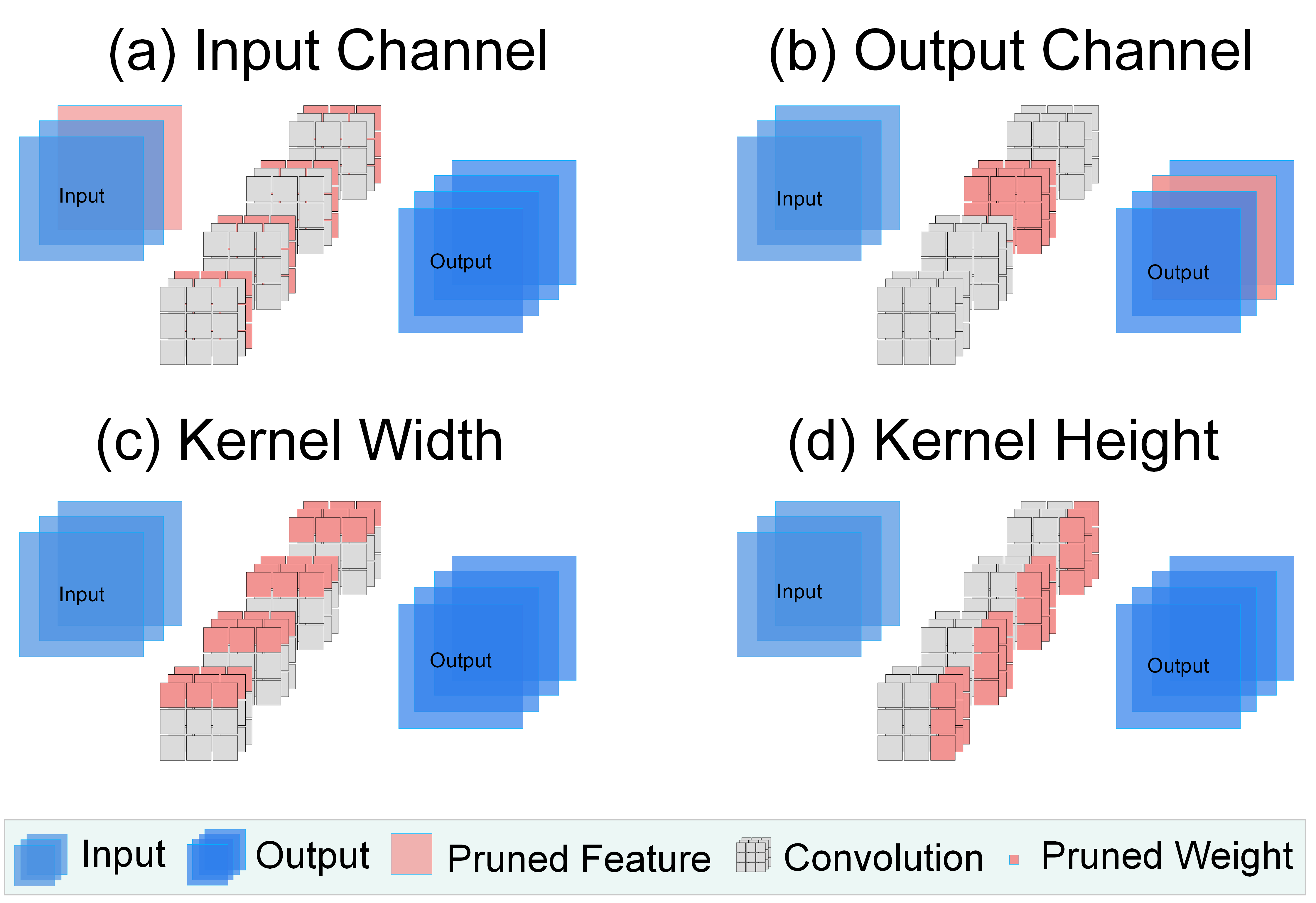}
    \caption{Illustration of Structural Pruning Impact between Feature Map and Convolution Weights}
    \label{fig:rw_pruning}
\end{center}\end{figure}

\begin{figure*}
    \centering
    \includegraphics[width=0.99\linewidth]{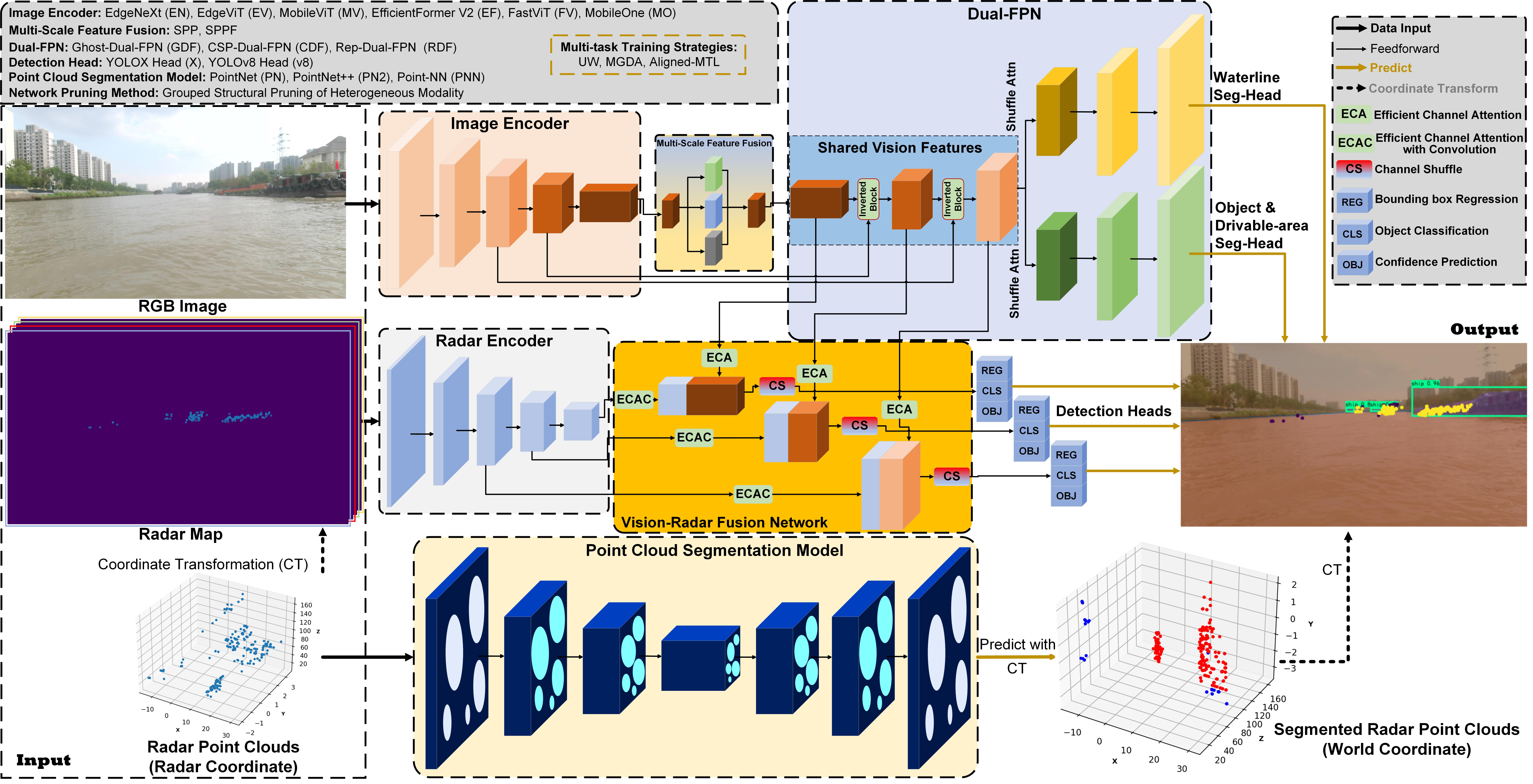}
    \caption{The architecture of Achelous++ framework. The prediction results include object bounding boxes, object masks, drivable-area (red), waterline (blue), object point clouds (non-blue), clutter point clouds (blue). The gray area in the upper-left corner shows the full and short name of each component. The gray area in the right center presents the meaning of symbols and part modules.}
    \label{fig:model}
\end{figure*}

In the case of a convolution module, we can perform pruning in four dimensions: input channels, output channels, and kernel sizes (height and width). As shown in Fig. \ref{fig:rw_pruning}, individually pruning the weights in these four dimensions allows us to directly reduce the size and quantity of the convolution kernels, leading to structural pruning \cite{li2016pruning}. Popular structured pruning algorithm \cite{he2023structured} contains Weight-Based \cite{li2016pruning}\cite{he2019filter}\cite{lin2022pruning}, Activation-Based \cite{lin2020hrank}\cite{hu2016network}\cite{luo2017thinet}\cite{yu2018nisp}, Regularization \cite{liu2017learning}\cite{chin2020towards}, and others \cite{lin2018accelerating}\cite{he2018amc}\cite{yu2022topology}. In this work, we focus on the output channel pruning with different layer-wise sparsity ratio.


\section{Achelous++}
\label{sec:framework}

\subsection{Overall Architecture}
Fig. \ref{fig:model} illustrates the comprehensive architecture of our Achelous++ framework. Achelous++ is characterized by a ternary-branch structure, featuring three distinct input streams, each corresponding to a dedicated encoder. This framework is conceived through the elegant amalgamation of encoder-decoder principles and the YOLO paradigm. Furthermore, we have made conscientious efforts to streamline the parallel and fragmented design aspects within Achelous++, aimed at enhancing inference speed while striking a balance between accuracy and efficiency.

\subsection{Input Data}

Although Achelous++ is a ternary-branch framework, there are two modal input data, RGB images captured by a monocular camera while radar point clouds obtained by a 4D mmWave radar in the radar coordinate. Each radar point cloud contains 3D coordinates $(x_r,y_r,z_r)$ under radar coordinate system and characteristics of the target range, compensated velocity, compensated height and reflected power. For the subsequent fusion operations between radar point clouds with RGB image, we transform the radar point clouds from the coordinate system of radar to camera plane, which is based on the extrinsic matrix between two sensors and the intrinsic matrix of camera. After obtaining the point clouds in the camera plane $(x_c, y_c)$, we concatenate these 2D radar point clouds of characteristics along the channel axis and get the radar map with four feature channels.

\subsection{Image Encoder}
Image encoder (backbone) is essential for image feature extraction and usually occupies the majority of the whole neural network. In response to the diverse requirements imposed by different computational devices, Achelous++ offers support for two distinct categories of backbones, namely, CNN-ViT hybrid networks \cite{maaz2022edgenext}\cite{pan2022edgevits}\cite{mehta2021mobilevit}\cite{li2023rethinking} and reparameterized networks \cite{vasu2023mobileone}\cite{vasu2023fastvit}. 

First and foremost, CNN-ViT hybrid networks represent a fusion of convolutional neural networks (CNN) with self-attention mechanisms. These networks excel in preserving data integrity, fending off adversarial attacks, and recognizing occluded objects in comparison to traditional CNN-based networks. Recent studies \cite{maaz2022edgenext}\cite{pan2022edgevits}\cite{mehta2021mobilevit}\cite{li2023rethinking} underscore that well-crafted CNN-ViT hybrid networks can achieve a speed comparable to the MobileNet series. However, it is important to note that CNN-ViT hybrid networks are the most intricate and resource-intensive among the three categories.

In addition to the hybrid networks, our image encoder also encompasses reparameterized networks, such as FastViT \cite{vasu2023fastvit} and MobileOne \cite{vasu2023mobileone}. These networks possess the capability to restructure the neural network from a parallel branch configuration to a more streamlined structure during the inference stage, resulting in a remarkable reduction in latency across a spectrum of computing devices.

Adhering to the conventional backbone paradigm, our image encoder is structured into five stages. Considering multiple tasks will increase the parallel task branches of the networks, to lower the complexity and reduce the inference latency, especially on performance-limit devices with various degrees, we empirically set three sizes of channels, which are S0, S1 and S2. The 5-stage channels of S0 are \{24, 32, 48, 96, 176\} while S1's are \{24, 32, 48, 120, 224\}. The largest channel sizes is S2, which are \{24, 32, 64, 144, 288\}.


\subsection{Multi-scale Feature Fusion}
To effectively integrate features from varying scales and receptive fields while mitigating the influence of object size on category recognition, we incorporate a spatial pooling pyramid (SPP) \cite{glenn_jocher_2020_4154370} into the image encoder. The utilization of SPP has demonstrated its efficacy in a wide array of vision-based tasks. However, it is noteworthy that traditional SPP involves multiple parallel branches, which can impede fast inference. To address this concern, we also introduce a slim variant known as SPPF \cite{glenn_jocher_2020_4154370} into our framework, simplifying the feature fusion process for enhanced computational efficiency during inference, which is also the default module in our models.

\subsection{Radar Map Encoder}

\begin{figure}
    \centering
    \includegraphics[width=0.99\linewidth]{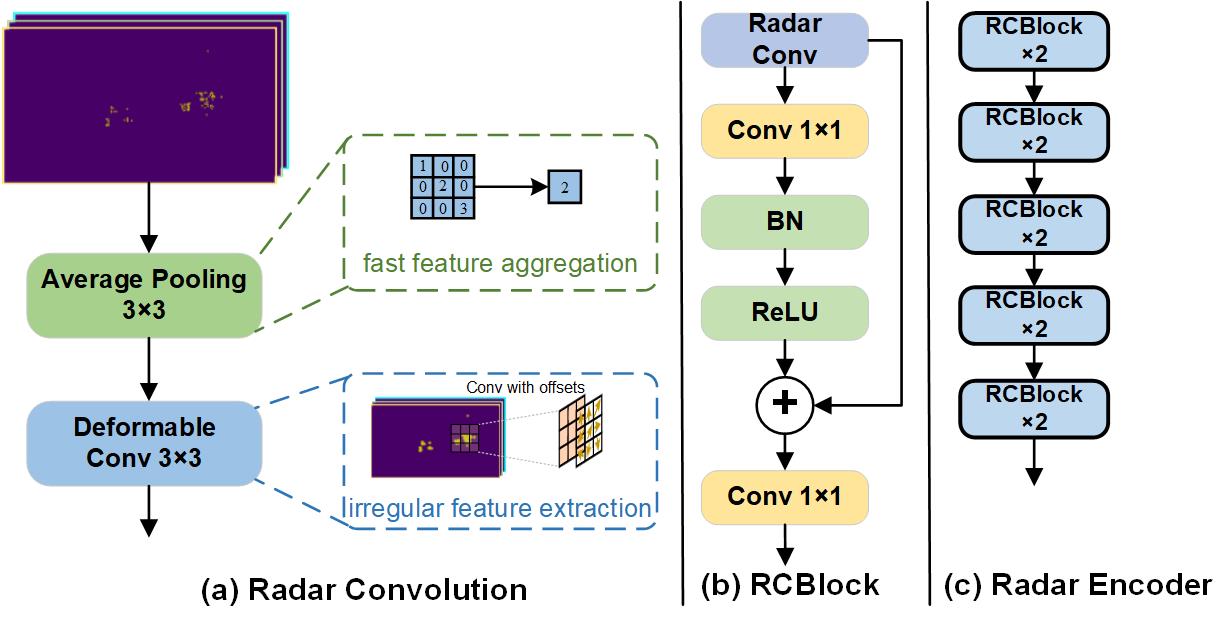}
    \caption{The radar encoder and its unit.}
    \label{fig:radar_conv}
\end{figure}

The Radar map encoder is tasked with the extraction and modeling of 2D radar point cloud features. It is widely understood that radar point clouds inherently exhibit sparsity and irregularity, after projecting from 3D sensor coordinate to 2D image plane, there are many large areas on the radar map without any value, only the areas where the target or clutter reflection points are located have irregular point cloud features. Therefore, we propose a novel and effective operator called Radar Convolution (RadarConv). RadarConv first applies a $3 \times 3$ average pooling to the radar map $f^r$ to aggregate the neighbourhood feature $f^r_a$ (Eq. \ref{eq:avg}). Compared with max-pooling, average pooling can maintain the neighborhood feature, rather than being disturbed by outliers. Moreover, the pooling operation is much faster than the convolution operation, especially on performance-limited devices. Secondly, a deformable convolution v2 \cite{zhu2019deformable} operator is to draw the irregular features $f^r_{da}$ (Eq. \ref{eq:dcn}).

\begin{align}
    & f^r_a = AvgPool(f^r), f^r_a \in R^{c \times h \times w} 
    \label{eq:avg} \\
    & f^r_{da} = \sum\limits_{k=1}^{K} w_k \cdot f^r_a(p+p_k + \Delta p_k) \cdot \Delta m_k, f^r_{da} \in R^{c \times h \times w}
    \label{eq:dcn}
\end{align}
where $K$ is the number of sampling offsets. Here, we set $K=9$. $p_k$ and $w_k$ indicate the pre-defined offset and the learnable weight at location $k$. $\Delta p_k$ and $\Delta m_k$ are learnable offset and modulated scalar at the position $k$. 

Upon RadarConv, we build the RCBlock as the unit of RCNet with five stages, whose number of channels per stage is a quarter of that of the image encoder. RCNet follows the paradigm of MobileNet \cite{howard2017mobilenets}. After the RadarConv extracts the irregular and sparse radar features, a $1 \times 1$ convolution is to weigh the spatial importance along the channel dimension. Then a batch normalization and a ReLU activation layer are attached. The output is added to the RadarConv result through a residual path. Finally, a $1 \times 1$ convolution is for feedforward. The whole process is presented in Eq. \ref{eq:rc1} and Eq. \ref{eq:rc2}.

\begin{align}
    & f^r_{rcb} = Conv_{1\times 1}(BN(ReLU(f^r_{da}))) + f^r_{da}
    \label{eq:rc1} \\
    & f^{r+1}_{rcb} = Conv_{1 \times 1} (f^r_{rcb})
    \label{eq:rc2}
\end{align}

\subsection{Dual Feature Pyramid Network}
Dual-FPN is employed to perform three distinct tasks: object semantic segmentation, drivable area segmentation, and waterline segmentation. However, the concurrent execution of multiple tasks means multiple parallel branches and a fragmented network design, which inevitably leads to a significant speed reduction and high complexity in the model.

\begin{figure}
    \centering
    \includegraphics[width=0.75\linewidth]{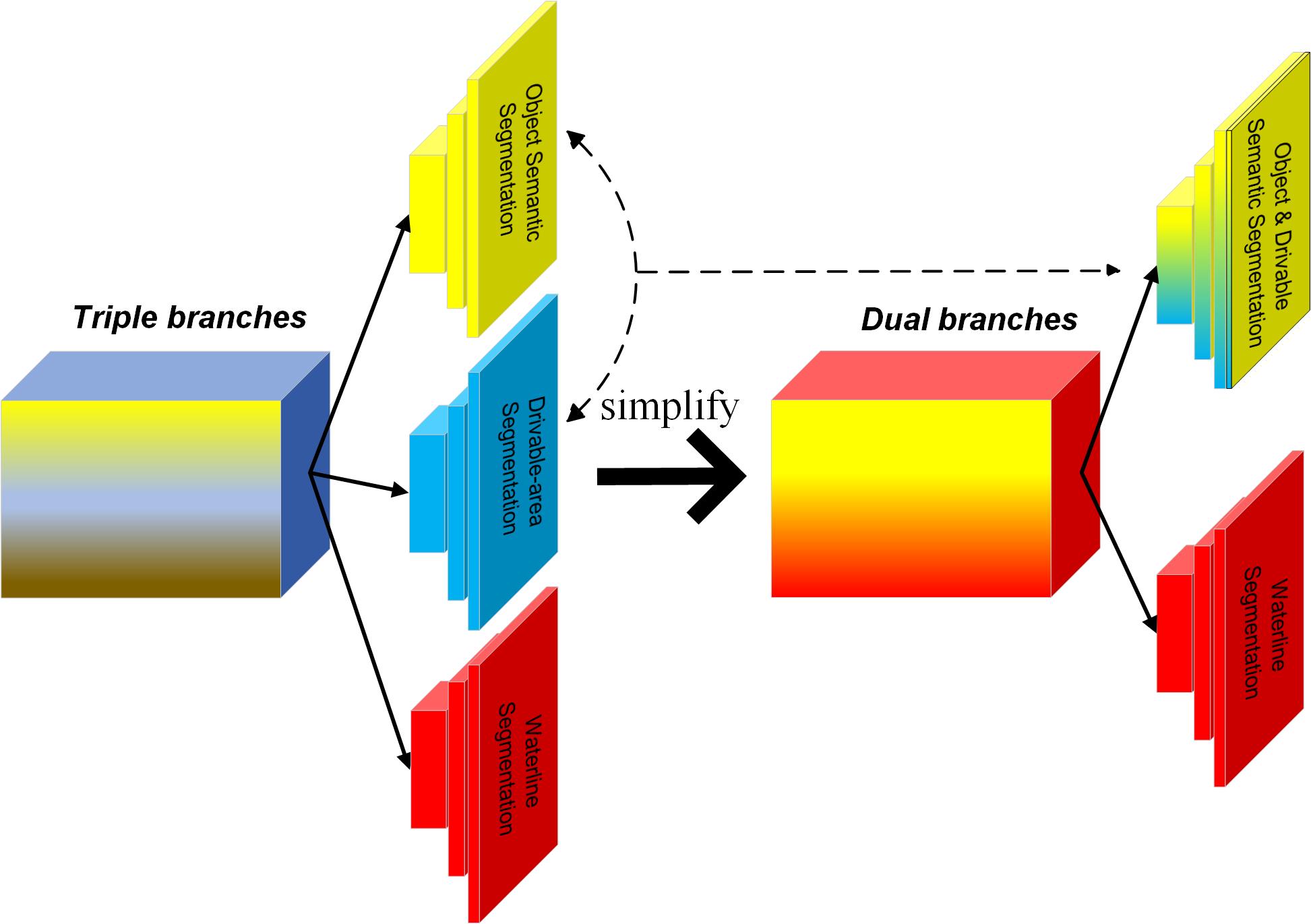}
    \caption{The simplification transformation from triple branches to dual branches.}
    \label{fig:simplify_triplet_dual}
\end{figure}

Thus, we contemplate the reduction of parallelized branches. In the case of water-surface objects, the drivable area consistently surrounds them, and They are each other's contextual features. However, with respect to the waterline, it generally appears on either side of the field of view, often adjacent to the drivable area. Yet, it consistently manifests as a long and narrow region in the image, exhibiting a significant disparity in scale compared to both the drivable area and the objects within the image. Moreover, it tends to be at a considerable distance from the objects, and its geometric characteristics differ substantially. To strike a balance between segmentation task speed and accuracy, as Fig. \ref{fig:simplify_triplet_dual} shows, we amalgamate the task of object segmentation with that of drivable area segmentation into one branch, while treating waterline segmentation as a distinct task branch. Detailedly, we concatenate one feature map on the high-resolution object feature segmentation head as the drivable-area prediction.

Our Achelous++ framework accommodates three lightweight FPN units: Ghost-DualFPN, CSP-DualFPN, and Rep-DualFPN. Ghost-DualFPN, based on the GhostNet architecture, excels in the reduction of feature redundancy. CSP-DualFPN leverages the well-established and efficient CSPDarkNet structure, widely recognized within the YOLO series networks. Additionally, Rep-DualFPN is built on the RepVGG block, offering complete reparameterization and an efficient FPN, with a specific focus on optimizing performance for high-performance computing devices.

In Dual-FPN, the first stage is the shared vision features, which contain various semantic information from the multi-scale network and skip connections from the image encoder. Assuming a feature map $f_s \in R^{c_1 \times h \times w}$ in shared vision features and another feature map $f_e \in R^{c_2 \times 2h \times 2w}$ from the image encoder, between $f_s$ and $f_{s+1}\in R^{c_3 \times 2h \times 2w}$, there is an inverted block to transform and upsample the feature. Eq. \ref{eq:up0}-\ref{eq:up3} present the whole process of inverted block. A $1 \times 1$ convolution is to adjust the channel dimension from $c_1$ to $c_2$ while a transform is also exerted on $f_e$, then we adopt interpolation with the scale factor of 2 to upsample the feature map and obtain $\tilde{f_s}$. Finally, we concatenate $\tilde{f_s}$ with $\hat{f_e}$ along the channel dimension and a unit module of Dual-FPN is to map the channel dimension from $2\times c_2$ to $c_3$.

\begin{align}
    & \hat{f_e} = Conv_{1\times 1}(f_e), \hat{f_e} \in R^{c_2 \times h \times w}
    \label{eq:up0} \\
    & \hat{f_s} = Conv_{1\times 1}(f_s), \hat{f_s} \in R^{c_2 \times h \times w}
    \label{eq:up1} \\
    & \tilde{f_s} = Upsample(\hat{f_s}, scale\_ factor=2), \tilde{f_s} \in R^{c_2 \times 2h \times 2w} 
    \label{eq:up2} \\
    & f_{s+1} = Unit(\hat{f_e} \oplus \tilde{f_s}), f_{s+1} \in R^{c_3 \times 2h \times 2w}
    \label{eq:up3}
\end{align}
where $\oplus$ denotes the concatenation operation.

Following the extraction of shared vision features, the Dual-FPN framework unfolds into two distinct branches, one specializing in waterline segmentation and the other in the segmentation of water surface targets and drivable areas. Firstly, instead of using skip-connection to feed multi-scale features into segmentation tasks, we separately exert a shuffle attention module \cite{zhang2021sa} to simultaneously reweigh the shared vision features, which includes much fewer embranchments and fragments and is much faster. After that, three-stage combinations of upsampling and feedforward operations are to restore the resolution of feature maps. Finally, two segmentation heads are attached to output the segmentation results.

\subsection{Vision-Radar Fusion Network}
Vision-Radar Fusion Network (VRFN) is a feature-level fusion network with FPN structure prepared for object detection. Here, we fuse shared vision features with multi-scale receptive fields with the radar maps of the last three stages with the same spatial shape. Before feeding the features of the two modalities into the VRFN, we first remeasure each feature map. For the radar feature map $f_r \in R^{c_r \times h \times w}$, we exert the Efficient Channel Attention with $1 \times 1$ Convolution (ECAC), which is to reweigh the channel and spatial features and alleviate the negative impact of radar clutter point features not in the target space. For the vision feature map $f_v \in R^{c_v \times h \times w}$, an ECA module is applied to perceive the channel weight for the pixel information in the same position. After that, we concatenate the feature maps of vision and radar, where radar maps without clutter features can provide the localization prior to vision feature by increasing channel dimension to locate targets and accelerate the convergence. To further reduce parameters, we apply Channel Shuffle (CS) to the fused feature. Channel shuffle reorganizes the channels of the original feature map, enhancing inter-channel information exchange while reducing inter-layer connection parameters, thereby reducing complexity and improving the model's generalization. The whole process is presented in Eq. \ref{eq:radar_weight}-\ref{eq:fuse_weight}.

\begin{align}
    & \hat{f_r} = Conv_{1\times 1}(ECA(f_r)), \hat{f_r} \in R^{c_r \times h \times w}
    \label{eq:radar_weight}, \\
    & \hat{f_v} = ECA(f_v), \hat{f_v} \in R^{c_v \times h \times w},
    \label{eq:vision_weight} \\
    & f_{vr} = CS(\hat{f_r} \oplus \hat{f_v}), \hat{f_{vr}} \in R^{(c_v+c_r) \times h \times w}, 
    \label{eq:fuse_weight}
\end{align}
where $\oplus$ denotes the concatenation operation.

\subsection{Detection Head}
Achelous++ supports two detection heads, including YOLOX decouple head \cite{ge2021yolox} and YOLOv8 decouple head \cite{yolov8_ultralytics}, which both take confidence prediction, bounding box regression and object classification as three separate tasks based on the multi-task learning idea. Accordingly, two label assignment strategies of SimOTA and Task Aligned Assigner in YOLOX and YOLOv8 are included.

\subsection{Radar Point Cloud Segmentation Model}
To avoid losing the accurate perception of the surrounding objects when the camera fails completely, we adopt the point cloud semantic segmentation model to predict a category for each point cloud under the radar coordinate system, which does not share any weights with the detection and segmentation networks. Our Achelous++ supports three models, including PointNet \cite{qi2017pointnet}, PointNet++ \cite{qi2017pointnet} and Point-NN \cite{zhang2023parameter}. PointNet and PointNet++ are two lightweight point cloud segmentation models with fast inference speeds. Due to the sparser nature of radar point cloud features compared to LiDAR and their limited high-level semantic information, as well as for the purpose of enhancing inference speed, we reduce the number of channels in PointNet and PointNet++ to one-third of their original values. Furthermore, Point-NN has no need to do this since it is a non-parametric zero-shot point cloud segmentation model without any training. 

\subsection{Output Data}
In alignment with the paradigm of YOLOP for road panoptic perception, Achelous++ collectively projects a series of predictions related to water environments onto the camera plane. These predictions encompass the detection of water surface obstacles, semantic segmentation of water surface obstacles, segmentation of drivable areas, segmentation of waterlines, and semantic segmentation of target point clouds. For point cloud coordinates, we transform the radar-based point cloud coordinates into camera-based coordinates through a coordinate system conversion. This transformation facilitates improved decision-making and understanding of obstacle positions, distances, and velocities by the model. By delivering results of panoramic perception, autonomous vessels can effectively perceive obstacles and pertinent semantic and physical information within their operational range, thereby furnishing comprehensive decision support for autonomous navigation.

\subsection{Multi-task Training Strategies}
Achelous++'s multi-task training strategy comprises two hierarchical levels, namely the fundamental loss functions and multi-task optimization strategies. In our task domain, we encompass loss functions for object detection, object and drivable area semantic segmentation, waterline segmentation, and point cloud semantic segmentation. Due to the lack of interrelation and backpropagation between the features used for point cloud semantic segmentation and other neural network components, the point cloud segmentation loss is computed independently and does not reside within the multi-task optimization strategy framework.

For fundamental loss functions, we employ a combination of Focal loss \cite{lin2017focal} and CIoU loss \cite{zheng2021enhancing} for object detection. In the context of all visual semantic segmentation tasks, we utilize a combination of Focal loss and Dice loss \cite{sudre2017generalised}. As for point cloud semantic segmentation, we employ Negative Log Likelihood (NLL) loss \cite{zhu2018negative}.

Concerning multi-task optimization strategies, in theory, our Achelous++ accommodates all multi-task optimization losses supported by LibMTL \cite{lin2023libmtl}. However, in this study, we optimize our multi-task model using three specific policies, namely, UW \cite{kendall2018multi}, MGDA \cite{sener2018multi}, and Aligned-MTL \cite{senushkin2023independent}. UW is founded on balancing multiple losses, while MGDA and Aligned-MTL are grounded in the balancing of parameter gradients across multiple task heads.

In summary, the goal of multi-task optimization in Achelous++ takes the form as illustrated in Equation \ref{eq:multi_loss}.

\begin{equation}
    min [ \sum \limits_{t=1}^{T} c^t L^t (\theta^{sh}, \theta^t) + L^{pc}(\theta^{pc}) ]
    \label{eq:multi_loss}
\end{equation}
where $T$ is the task number with shared parameters $\theta^{sh}$. $\theta^t$ is the parameter of each individual task and $c^t$ is the static or dynamic learnable weight. Here, $T=3$, where it originally denotes to four tasks with shared parameter space, including object detection, object semantic segmentation, drivable-area segmentation and waterline segmentation, since we combine the object segmentation and drivable-area segmentation as one task, the task number in the multi-task loss becomes three. Moreover, since the feature of point cloud semantic segmentation $\theta^{pc}$ does not share the feature parameter with other tasks, we exclude the loss of point cloud semantic segmentation $L^{pc}$ from the multi-task balanced loss function and add it separately.

\subsection{Heterogeneous Modalities Pruning} \label{sec:prune}
The heterogeneous modalities structural pruning specifically focuses on manipulating the input and output of modules (convolution, linear, bn, etc.) within the Achelous++. The basic pruning elements are the number of filters (output channel count), and the input channel count serves as a complementary consequence of pruning the previous layer's filters. 

\subsubsection{Nodes and Groups}
Inspired by DepGraph \cite{fang2023depgraph}, we can construct a relationship graph among all modules in the model based on the backward propagation order. In this study, we categorize all nodes into the following types:
\begin{itemize}
    \item \textbf{In-Out} These nodes have both input and output channels that can be pruned, such as convolution layers and linear layers.
    \item \textbf{Out-Out} These nodes have only one channel that can be pruned, and the pruning result is solely determined by the preceding layer's nodes, like normalization layers and depth-wise separable convolutions.
    \item \textbf{In-In} This is a special type of node, including operations like add and sub, which do not have pruning weights themselves. However, they must ensure that the pruning weight indices from previous layers are consistent, which is crucial in the grouping process discussed in the next section.
    \item \textbf{Remap} This mainly consists of special nodes like concat and split, which play a crucial role in aggregating or separating indices.
    \item \textbf{Reshape} These nodes are responsible for reshaping feature maps, such as flatten, permute, expand, and transpose operations.
    \item \textbf{Dummy} This category primarily includes output and input nodes that do not have pruning parameters, and they require that connected channels cannot be pruned.
    \item \textbf{Custom} These are special nodes that we introduce when the pruning framework cannot handle certain module types. Custom nodes are needed to be assigned to one of the above node types or can be used as an additional solution when certain module attributes cannot be easily obtained but play a vital role in inference. For example, in ShuffleNet, learnable parameters cannot be easily obtained in Achelous++, but they are also needed to be pruned as well.
\end{itemize}

\begin{figure}[ht]\begin{center}
    \includegraphics[width=0.97\linewidth]{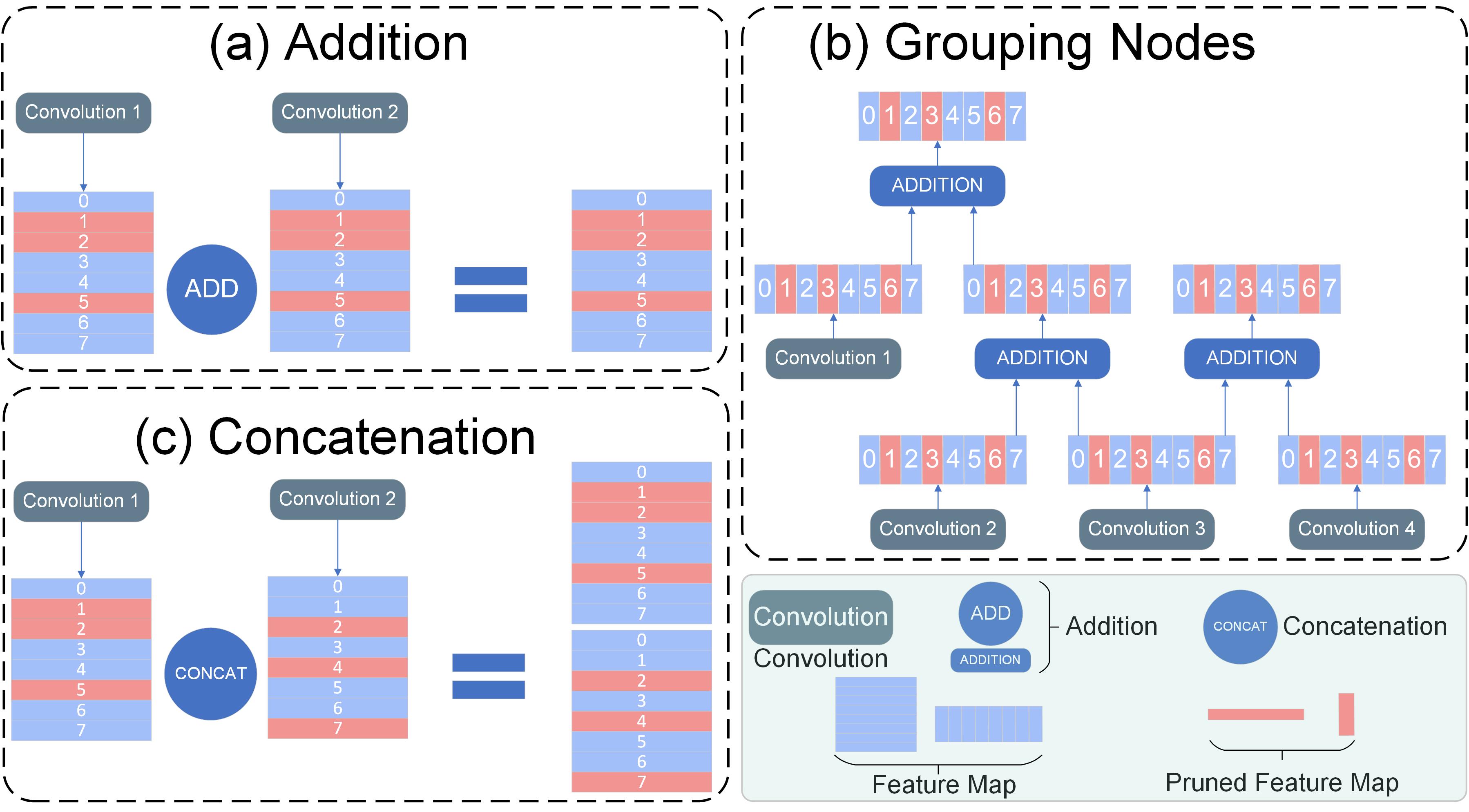}
    \caption{Pruning-index-based grouping strategies of VRFN in Achelous++. There are two different type of nodes (a) and (b) that affect the pruning index. The grouping strategy is searching all connected nodes in (b).}
    \label{fig:md_pruning_sample}
\end{center}\end{figure} 

Once the relationship network is constructed, modules within the same layer need to be grouped together. The grouping criteria are as follows: ``Modules within the same group have consistent output channels, and either the current or the next layer contains the same input-output node. If the next layer contains an output-output node, the search continues for the next node." There are two different type of nodes that affect the pruning indices, shown in Fig.\ref{fig:md_pruning_sample} (a) and (c). If an \textbf{add node} contains two convolution layers, both of these convolution layers will undergo filter pruning simultaneously, and their pruning indices will be identical, shown in Fig.\ref{fig:md_pruning_sample} (a). When calculating the pruning scores for the filter dimensions, both layers will be considered together. In contrast, as shown in Fig.\ref{fig:md_pruning_sample} (c), the \textbf{concat node} plays a similar role but does not require the pruning indices of the preceding layer's nodes to be identical. To conclude, as depicted in Fig.\ref{fig:md_pruning_sample} (b), we utilize the add node as criteria for searching groups. When an add node connects multiple convolutions and add nodes, all the nodes are found, and the entire group is considered as one group. 

\subsubsection{ERK Ratio Pruning} 

To preserve more weights, we select the lowest ERK sparsity in one group. Moreover, due to our utilization of PaI strategy, when it comes to pruning the output filter count, we employ random indices for the process. ERK ratio for convolutional layer \cite{evci2020rigging} and linear layer \cite{mocanu2018scalable} could be expressed by:

\begin{align}
(\text{conv})\quad & \rho_{ERK} = \rho_{global}\left(1-\frac{n^{l-1}+n^{l}+k_1^{l}+k_2^{l}}{n^{l-1}\times n^{l}\times k_1^{l}\times k_2^{l}}\right) \label{eq:md_pruning_erk1}\\
(\text{linear})\quad & \rho_{ERK} = \rho_{global}\left(1-\frac{n^{l-1}+n^{l}}{n^{l-1}\times n^{l}}\right), \label{eq:md_pruning_erk2}
\end{align}
where the $\rho_{global}$ is the predefined global setting of pruning sparsity ratio, $n$ is the channel or filter count, $k$ is the convolution kernel size. In addition, the pruning sparsity ratio $\rho$ describe the pruned network weight ratio, rather than the remained network weight ratio.

\subsubsection{Heterogeneous Aware SynFlow Pruning}
The tags of two different input sources are assigned to each node in Achelous++ pruning graph, which divide the nodes into three different manner: radar, vision, and both. By checking the tag of nodes and its related nodes, we could simply find the fusion stage in Achelous++. Then, we collect the incoming nodes of fusion stage belonging to each input source.

SynFlow \cite{tanaka2020pruning} unstructural pruning algorithm introduced and proved the neuron-wise conservation of synaptic saliency in a model. Based on the conservation, we evaluate the weight importance of two modalities in pre-fusion nodes, which directly contributes to fusion stages. HA-SynFlow algorithm could be expressed as:

\begin{align}
    \mathbb{S}_{SF} &= \frac{\partial \mathbb{R}}{\partial \theta} \odot \theta \label{eq:md_HA1}\\
    \mathbb{S}_{HASF,m} &= \frac{1}{N_{m}}\sum^{N_m}_{n=1}\left| \mathbb{S}_{SF,\theta_{i}} \right|, \label{eq:md_HA2}
\end{align}
where $\theta$ is the weight for a module, $\mathbb{R}$ is $\mathbb{1}^T\left(\prod_{l=1}^L\left|\theta^{[l]}\right|\right) \mathbb{1}$, $N_{m}$ is the weight parameter count for the modality $m$. Then we could obtain the pruning sparsity based on the modality-wise score:

\begin{align}
    \bar{\mathbb{S}_{HASF}} &= \frac{1}{N_M}\sum_{m=1}^{M} \mathbb{S}_{HASF,m} \label{eq:md_sparsity1}\\
    \rho_{m} &= \rho_{global} + (1-\rho_{global}) \cdot \log \left(\frac{\bar{\mathbb{S}_{HASF}}}{\mathbb{S}_{HASF,m}}\right) \label{eq:md_sparsity2}\\
    \rho_{M^\prime} &= \frac{1}{N_{M^\prime}}\sum_{m^\prime=1}^{M^\prime} \rho_{m^\prime}, \label{eq:md_sparsity3}
\end{align}
where $N_{M}$ represents the number of modalities, and the $M^\prime$ is the modules containing more than one modalities.


\section{Experiments}
\label{sec:experiments}

\subsection{Dataset}
We train and evaluate our Achelous++ on WaterScenes \cite{yao2023waterscenes} dataset, which is currently the largest multi-task water-surface perception benchmark based on camera and 4D mmWave radar, including various urban and countryside scenarios. WaterScenes contains 54,120 frames and 202,807 objects, including pier, buoy, sailor, ship, boat, vessel and kayak, a total of seven categories for object detection and object semantic segmentation. WaterScenes also has pixel-wise annotations of drivable area and waterline for segmentation of drivable area and waterline. Furthermore, WaterScenes has the point-level category annotation of 4D radar point cloud, including the object categories and clutter, a total of eight categories for point cloud semantic segmentation.

\subsection{Settings of Training and Evaluation}
During the scratch training, we resize the image and radar map as 320 $\times$ 320 pixels and set the batch size as 32. We train all models for 100 epochs with an initial learning rate of 0.03. We adopt a cosine learning rate scheduler and AdamW optimizer. As a multi-task learning framework, Achelous++ also supports LibMTL \cite{lin2023libmtl}, a comprehensive multi-task learning strategy library to optimize our Achelous++. Here, to accelerate training speed, we adopt UW \cite{kendall2018multi} as the multi-task learning strategy to train these models here. Furthermore, we adopt Exponential Moving Average (EMA) to alleviate the overfitting of the model and improve the generalization and stability. To save the GPU memory, the training framework also supports mixed precision training. Moreover, our Achelous++ is compatible with both Windows and Linux operating systems. Accordingly, Data Parallel (DP) and Distributed Data Parallel (DDP) are supported to accelerate the training on Windows and Linux platforms individually. In this paper, we train all models on two RTX A4000 GPUs.

During the inference, we adopt abundant metrics to evaluate the performances of models from various perspectives. For the object detection, AP$_\text{50}$, mAP$_\text{50-95}$ and AR$_\text{50}$ are for the evaluation comprehensively. For the object semantic segmentation, drivable area segmentation and waterline segmentation, mIoU is selected to evaluate the performances. Semantic segmentation of radar point cloud adopts mIoU as the metric. Moreover, to evaluate the inference speed comprehensively, we select NVIDIA Jetson Xavier NX, NVIDIA AGX Orin, GTX 1650 and RTX 3090ti as the inference platforms with different GPU computing architectures. Last but not least, we evaluate the Energy consumed Per Sample (EPS) and AVerage Power (AVP) of all models on NVIDIA AGX Orin and GTX 1650 separately.

\begin{align}
    \textit{EPS} &= \frac{1}{2N_{s}} \cdot \left(\sum ^{N_t}_{n=0} \tau_{n} \cdot (P_{n}+P_{n+1}-2P^{\prime})\right) \\
    \textit{AVP} &= N_{s} \cdot \textit{EPS} \cdot \left(\sum ^{N_{t}}_{n=0} \tau_{n}\right)^{-1}
\end{align}
where the $N_t$ and $N_s$ are the discrete time sample point and number of evaluation samples, the $\tau_n$ represents the time interval of two time sample point, the $P^{\prime}$ is the background power of device, $P_n$ and $P_{n+1}$ are instantaneous power for time $n$ and $n+1$. Similar to Femtodet \cite{tu2023femtodet}, the power is calculated with the estimation API from hardware provider, and the background power $P^{\prime}$ is calculated before inference. We perform warmup strategy at the beginning of evaluation, which ensure representation of real-world scenarios.



\subsection{Overall Performances of Various State-of-the-art Models on Different Tasks}
\begin{table*}
\center
\footnotesize
\setlength\tabcolsep{1.8pt}
\caption{Overall Comparison of Various Models on Test Set of WaterScenes.}
  \label{tab:prediction_results}
\begin{tabular}{l|cc|cc|ccc|cc|c|c|cccc}
\toprule
\multicolumn{1}{c}{\multirow{2}[2]{*}{\textbf{Models}}} &
\multicolumn{1}{c}{\multirow{2}[2]{*}{\textbf{Sensors}}} &
\multicolumn{1}{c}{\multirow{2}[2]{*}{\textbf{TN$^1$}}} & 
\multicolumn{1}{c}{\multirow{2}[2]{*}{\textbf{Params(M)}}} & 
\multicolumn{1}{c}{\multirow{2}[2]{*}{\textbf{FLOPs(G)}}} & 
\multicolumn{3}{c}{\bf{OD$^2$}} &
\multicolumn{2}{c}{\bf{SS$^3$}} & 
\multicolumn{1}{c}{\bf{WS$^4$}} & 
\multicolumn{1}{c}{\bf{PC-SS$^5$}} & 
\multicolumn{1}{c}{\multirow{2}[2]{*}{\textbf{FPS$_{\text{J}}$$^9$}}} &
\multicolumn{1}{c}{\multirow{2}[2]{*}{\textbf{FPS$_{\text{O}}$$^{10}$}}} &
\multicolumn{1}{c}{\multirow{2}[2]{*}{\textbf{FPS$_{\text{G}}$$^{11}$}}} &
\multicolumn{1}{c}{\multirow{2}[2]{*}{\textbf{FPS$_{\text{R}}$$^{12}$}}}
  \\ \cmidrule(lr){6-8}\cmidrule(lr){9-10}\cmidrule(lr){11-11}\cmidrule(lr){12-12}
\multicolumn{5}{c}{} & \bf{mAP$_{50\text{-}95}$} & \bf{mAP$_{50}$} & \bf{AR$_{50\text{-}95}$} & \bf{mIoU$_{\text{t}}$$^6$} & \bf{mIoU$_{\text{d}}$$^7$} & \bf{mIoU$_{\text{w}}$$^8$} & \bf{mIoU}     
\\\midrule
\multicolumn{16}{c}{\texttt{Single-Task Models}} \\
\midrule
\multicolumn{16}{c}{Object Detection} \\
\midrule
YOLOv8-Nano \cite{yolov8_ultralytics} & C$^{13}$ & 1 & \textbf{3.01} & \textbf{2.05} & \textbf{41.9} & \textbf{71.8} & 44.0 & - & - & - & - & 29.8 & 51.9 & 117.8 & 135.7  \\
YOLOv7-Tiny \cite{wang2023yolov7} & C &  1 & 6.03 & 33.3 & 37.3 & 65.9 & 43.7 & - & - & - & - & 36.7 & 81.1 & 241.3 & 289.5 \\
YOLOX-Tiny \cite{ge2021yolox} & C &  1 & 5.04 & 3.79 & 39.4 & 68.0 & 43.0 & - & - & - & - & 33.6 & 77.6 & 132.6 & 170.1 \\
YOLOv4-Tiny \cite{bochkovskiy2020yolov4} & C &  1 & 5.89 & 4.04 & 13.1 & 36.3 & 20.2 & - & - & - & - & \textbf{114.6} & \textbf{175.8} & \textbf{281.2} & \textbf{352.2} \\
CRFNet \cite{nobis2019deep} & \textbf{C+R} & 1 & 23.54 & - & 41.8 & 71.2 & \textbf{44.5} & - & - & - & - & - & - & - & - \\
\midrule
\multicolumn{16}{c}{Semantic Segmentation} \\
\midrule
Segformer-B0 \cite{xie2021segformer} & C &  1 & 3.71 & 5.29 & - & - & - & \textbf{73.5} & \textbf{99.4} & \textbf{72.9} & - & 41.6 & 46.3 & 89.3 &  184.7\\
PSPNet \cite{zhao2017pyramid} & C &  1 & \textbf{2.38} & \textbf{2.30} & - & - & - & 69.4 & 99.0 & 69.7 & - & \textbf{61.2} & \textbf{61.2} & \textbf{149.3} &  \textbf{273.1} \\
\midrule
\multicolumn{16}{c}{Point Cloud Semantic Segmentation} \\
\midrule
PointNet \cite{qi2017pointnet} & R$^{14}$ &  1 & 3.53 & \textbf{1.19} & - &- &- &- &- &- & 59.9 & \textbf{97.0} & \textbf{112.3} & \textbf{203.2} &  \textbf{568.4} \\
PointNet++ \cite{qi2017pointnet++} & R &  1 & 1.88 & 2.63 & - &- &- &- &- &- & \textbf{60.7} & 72.8 & 100.6 & 187.4 &  401.2 \\
Point-NN \cite{zhang2023parameter} & R &  1 & \textbf{0.0} & 1.98 & - &- &- &- &- &- & \textbf{52.7} & 51.3 & 73.5 & 142.5 & 321.7  \\
\midrule
\midrule
\multicolumn{16}{c}{\texttt{Multi-Task Models}}  \\
\midrule
\multicolumn{16}{c}{Panoptic Perception} \\
\midrule
YOLOP \cite{wu2022yolop} & C & 3 & 7.90 & 18.60 & 37.9 & 68.9 & 43.5 & - & 99.0 & \textbf{74.9} & - & \textbf{16.2} & \textbf{23.1} & \textbf{79.3} &  \textbf{86.5} \\
HybridNets \cite{vu2022hybridnets} & C & 3 & 12.83 & 15.60 & 39.1 & 69.8 & 44.2 & - & 98.8 & 71.5 & - & 6.04 & 7.8 & 17.4 & 30.8 \\
Efficient-VRNet-N \cite{guan2023efficient} & \textbf{C+R} & 3 & \textbf{4.10} & \textbf{3.21} & \textbf{42.0} & \textbf{73.5} & \textbf{44.2} & \textbf{72.1} & \textbf{99.3} & - & - & - & 8.0 & 28.3 & -\\
\midrule
\multicolumn{16}{c}{\textbf{Ours (CNN-ViT Hybrid Backbones with Normal Neural Network Structures)}} \\
\midrule
\textbf{\textit{S0}} \\
\midrule
\textbf{EN-CDF-X-PN} & \textbf{C+R} & \textbf{5} & 3.59 & 5.38 & 37.2 & 66.3 & 43.1 & 68.1 & 98.8 & 69.4 & 57.1 & 17.5 & 22.0 & 52.2 & 72.8  \\
\textbf{EN-GDF-X-PN} & \textbf{C+R} & \textbf{5} & 3.55 & 2.76 & 37.5 & 66.9 & 44.6 & 69.1 & 99.0 & 69.3 & 57.8 & \textbf{17.8} & \textbf{22.3} & 59.7 & 72.5 \\
\textbf{EN-CDF-v8-PN} & \textbf{C+R} & \textbf{5} & 3.51 & 5.29 & 37.4 & 66.1 & 43.3 & 68.1 & 98.8 & 69.4 & 57.1 & 17.2 & 21.6 & 51.7 & \textbf{73.1}  \\
\textbf{EN-GDF-v8-PN} & \textbf{C+R} & \textbf{5} & 3.51 & \textbf{2.69} & 37.4 & 67.1 & 44.5 & 69.1 & 99.0 & 69.3 & 57.8 & 17.6 & 22.2 & 58.7 & 72.3 \\
\textbf{EN-CDF-X-PN2} & \textbf{C+R} & \textbf{5} & 3.69 & 5.42 & 37.3 & 66.3 & 43.0 & 68.4 & 99.0 & 68.9 & \textbf{60.2} & 15.2 & 20.9 & 49.8 & 71.6 \\
\textbf{EN-GDF-X-PN2} & \textbf{C+R} & \textbf{5} & 3.64 & 2.84 & 37.7 & 68.1 & 45.0 & 67.2 & 99.2 & 67.3 & 59.6 & 14.8 & 21.8 & 56.8 & 71.0  \\
\textbf{EF-GDF-X-PN} & \textbf{C+R} & \textbf{5} & 5.48 & 3.41 & 37.4 & 66.5 & 43.4 & 68.7 & \textbf{99.6} & 66.6 & 59.4 & 17.3 & 20.3 & 59.9 & 62.2 \\
\textbf{EV-GDF-X-PN} & \textbf{C+R} & \textbf{5} & 3.79 & 2.89 & 38.8 & 67.3 & 42.3 & 69.8 & \textbf{99.6} & \textbf{70.6} & 58.0 & 16.4 & 21.7 & \textbf{66.6} & 69.5\\
\textbf{MV-GDF-X-PN} & \textbf{C+R} & \textbf{5} & \textbf{3.49} & 3.04 & \textbf{41.5} & \textbf{71.3} & \textbf{45.6} & \textbf{70.6} & 99.5 & 68.8 & 58.9 & 16.0 & 20.2 & 60.5 & 64.7\\ 
\midrule
\textbf{\textit{S1}} \\
\midrule
\textbf{EN-GDF-X-PN} & \textbf{C+R} & \textbf{5} & 5.18 & 3.66 & 41.3 & 70.8 & 45.5 & 67.4 & 99.4 & \textbf{69.3} & 58.8 & 16.6 & 20.8 & \textbf{55.5} & \textbf{69.0} \\
\textbf{EF-GDF-X-PN} & \textbf{C+R} & \textbf{5} & 8.07 & 4.52 & 40.0 & 70.2 & 43.8 & 68.2 & 99.3 & 68.7 & 58.2 & 16.6 & 18.5 & 52.2 & 58.1 \\
\textbf{EV-GDF-X-PN} & \textbf{C+R} & \textbf{5} & \textbf{4.14} & \textbf{3.16} & 41.0 & 70.7 & 45.9 & 70.1 & 99.4 & 67.9 & \textbf{59.2} & \textbf{16.7} & \textbf{21.1} & 50.6 & 68.9   \\
\textbf{MV-GDF-X-PN} & \textbf{C+R} & \textbf{5} & 4.67 & 4.29 & \textbf{43.1} & \textbf{75.8} & \textbf{47.2} & \textbf{73.2} & \textbf{99.5} & 69.2 & 59.1 & 15.8 & 20.0  & 51.8 & 64.2\\
\midrule
\textbf{\textit{S2}} \\
\midrule
\textbf{EN-GDF-X-PN} & \textbf{C+R} & \textbf{5} & \textbf{6.90} & \textbf{4.59} & 40.8 & 70.9 & 44.4 & 69.6 & 99.3 & 71.1 & \textbf{59.0} & \textbf{16.1} & \textbf{20.7}  & \textbf{50.3} & \textbf{68.2} \\
\textbf{EF-GDF-X-PN} & \textbf{C+R} & \textbf{5} & 14.64 & 7.13 & 40.5 & 70.8 & 44.5 & 70.3 & 99.1 & \textbf{71.7} & 58.4 & 13.5 & 15.5 & 40.5 & 58.1 \\
\textbf{EV-GDF-X-PN} & \textbf{C+R} & \textbf{5} & 8.28 & 5.19 & 40.3 & 69.7 & 43.8 & \textbf{74.1} & 99.5 & 67.9 & 58.3 & 14.7 & 17.9 & 46.6 & 56.1\\
\textbf{MV-GDF-X-PN} & \textbf{C+R} & \textbf{5} & 7.18 & 6.02 & \textbf{45.0} & \textbf{79.4} & \textbf{48.8} & 73.8 & \textbf{99.6} & 70.8 & 58.5 & 15.6 & 19.6 & 48.8 & 63.8\\
\midrule
\multicolumn{16}{c}{\textbf{Ours (Reparameterization Structures)}} \\
\midrule
\textbf{\textit{S0}} \\
\midrule
\textbf{MO-RDF-X-PN} & \textbf{C+R} & \textbf{5} & \textbf{2.74} & \textbf{8.49} & \textbf{39.8} & 72.1 & \textbf{44.4} & 70.4 & \textbf{99.1} & \textbf{65.3} & 59.1 & \textbf{23.1} & \textbf{28.6} & \textbf{56.1} &  \textbf{96.3} \\
\textbf{FV-RDF-X-PN} & \textbf{C+R} & \textbf{5} & 3.49 & 9.01 & 38.8 & \textbf{73.3} & 43.6 & \textbf{72.1} & 98.8 & 64.8 & \textbf{60.1} & 21.8 & 26.0 & 52.6 &  87.8 \\
\midrule
\textbf{\textit{S1}} \\
\midrule
\textbf{MO-RDF-X-PN} & \textbf{C+R} & \textbf{5} & \textbf{2.98} & \textbf{8.64} & \textbf{40.0} & 72.8 & \textbf{44.4} & 70.7 & 99.2 & 65.5 & 59.8 & \textbf{22.2} & \textbf{28.0} & \textbf{55.3} &  \textbf{95.2} \\
\textbf{FV-RDF-X-PN} & \textbf{C+R} & \textbf{5} & 4.33 & 9.45 & 39.5 & \textbf{74.8} & 44.3 & \textbf{73.9} & \textbf{99.3} & \textbf{65.7} & \textbf{60.0} & 20.6 & 24.9 & 50.7 &  84.9 \\
\midrule
\textbf{\textit{S2}} \\
\midrule
\textbf{MO-RDF-X-PN} & \textbf{C+R} & \textbf{5} & \textbf{3.42} & \textbf{9.49} & \textbf{42.1} & 75.1 & \textbf{44.6} & 71.9 & \textbf{99.5} & 67.6 & 59.7 & \textbf{21.6} & \textbf{27.9} & \textbf{53.1} & \textbf{93.1} \\
\textbf{FV-RDF-X-PN} & \textbf{C+R} & \textbf{5} & 6.10 & 10.98 & 41.2 & \textbf{75.9} & 44.3 & \textbf{74.8} & 99.4 & \textbf{67.8} & \textbf{60.0} & 19.3 & 24.5 & 47.7 &  84.2 \\

\bottomrule
\end{tabular}
\\
\vspace{1mm}
\scriptsize{1. \textbf{TN}: task number 2. \textbf{OD}: object detection 3. \textbf{SS}: semantic segmentation 4. \textbf{WS}: waterline segmentation 5. \textbf{PC-SS}: point cloud semantic segmentation 6. mIoU$_{\text{t}}$: mIoU of targets 7. mIoU$_{\text{d}}$: mIoU of drivable area 8. mIoU$_{\text{w}}$: mIoU of waterline 9. FPS$_{\text{J}}$: FPS on NVIDIA Jetson Xavier 10. FPS$_{\text{O}}$: FPS on NVIDIA AGX Orin 11. FPS$_{\text{G}}$: FPS on NVIDIA GTX 1650. 12. FPS$_{\text{R}}$: FPS on NVIDIA RTX 3090ti. 13. C: camera 14. R: radar.}
\end{table*}

Table \ref{tab:prediction_results} presents comprehensive experimental results on the test set of WaterScenes. We categorize all participating models in the evaluation into two major groups: single-task models and multi-task models. To ensure a fair comparison, we select state-of-the-art and classic models with comparable parameter counts. The single-task visual object detection models included YOLOv8-Nano, YOLOv7-Tiny, YOLOX-Tiny, YOLOv4-Tiny, and CRFNet, a classical radar-visual fusion detection model. For visual segmentation tasks, we utilize the hybrid network Segformer-B0 and the pure convolutional network PSPNet. In the context of radar point cloud semantic segmentation, we employ lightweight models such as PointNet, PointNet++, and the parameterless model Point-NN. In contrast, among the multi-task models, we opt for prominent models such as YOLOP for visual panoramic perception, HybridNets, and the radar-visual asymmetric fusion multi-task network Efficient-VRNet-N.

Within our Achelous++ framework, the models are categorized into two primary groups. One group employs a hybrid backbone based on CNN-ViT architecture, and these models predominantly utilize Ghost-DualFPN (GDF) as the foundational module for segmentation tasks. The other group features fully reparameterized structures and relies entirely on reparameterization. Both of these categories extensively experiment with the use of YOLOX decoupled detection heads (X) and PointNet (PN) as foundational architectural components. 

It is evident that the Achelous++ series models have achieved highly advanced performance overall. Furthermore, compared to the models mentioned above, Achelous++ supports two modal inputs and is capable of accomplishing five tasks with fewer parameters and FLOPs.

For object detection, the purely visual YOLOv8-Nano, fusion-based CRFNet, and Efficient-VRNet-N models have all delivered quite comparable results, with mAP$_{\text{50-95}}$ ranging from 41.8 to 42.0. Fusion models clearly outperform in terms of recall, indicating that fusion can, to a certain extent, reduce false negatives. The Achelous++ series models achieve the highest mAP$_{\text{50-95}}$ of 45.0. Notably, models using MobileViT as the backbone perform exceptionally well in the S0, S1, and S2 size series, seemingly validating the efficacy of native multi-head self-attention in reducing false negatives. Additionally, we observe that YOLOv4 and YOLOv7, which utilize COCO dataset anchor priors, underperformed, whereas anchor-free models like YOLOv8 and YOLOX excel.

In the context of visual semantic segmentation, we observe that FV-RDF-X-PN (S2) achieves the state-of-the-art mIoU (74.8) for water surface object segmentation, with the highest values for S1 and S0 models also attained by FV-RDF-X-PN. Among the single-task models, Segformer-B0 achieves a mIoU of 73.5, which is lower than FV-RDF-X-PN. As for drivable area segmentation, all models perform remarkably close, ranging between 98.8 and 99.6. However, it is undeniable that our Achelous++ still achieves the state-of-the-art mIoU (99.6). In waterline segmentation, YOLOP achieve an mIoU of 74.9, outperforming our EF-GDF-X-PN (71.7), while Segformer-B0 also achieves an mIoU of 72.9.

For radar point cloud semantic segmentation, we observe that reducing the number of channels does not adversely affect the results. This, to some extent, demonstrates that when dealing with sparse radar point clouds, lightweight models can be effectively employed without sacrificing the quality of results.

\subsection{Analysis of Inference Speed}
In terms of model inference speed, we conduct a rigorous comparison of our Achelous++ framework against single-task and other single-modal multi-task models. To comprehensively evaluate the Achelous++ framework, we utilize four different test devices, including two edge devices and two host GPUs, namely Jetson Xavier NX (Volta architecture), Jetson AGX Orin (Ampere architecture), GTX 1650 (Turing architecture), and RTX 3090ti (Ampere architecture). 

In Table \ref{tab:prediction_results}, it is evident that single-task models exhibit notably faster speeds compared to other multi-task models. Regarding multi-task models, we observe that our Achelous++ achieves faster speeds compared to YOLOP, HybridNets, and Efficient-VRNet-N. Notably, EN-GDF-X-PN (S0) achieves 17.8 FPS on Xavier and 22.3 FPS on Orin, while the largest size EN-GDP-X-PN (S2) also reaches 16.1 and 20.7 FPS on Xavier and Orin, respectively. For reparameterized models, MO-RDF-X-PN (S0) achieves 23.1 and 28.0 FPS on both edge devices. On host GPUs, all models under the Achelous++ framework achieve speeds exceeding 40 FPS. In conclusion, Achelous++ offers real-time perception capabilities for USV autonomous navigation on the water surface.

Of greater significance, we have observed that reparameterization plays a crucial role in accelerating model inference. As evidenced by their performance across diverse tasks and the overall inference speeds, MobileOne, FastViT, and RepVGG excel in converting parallel branches into equivalent single-stream structures during inference, resulting in a substantial boost in inference speed while maintaining task accuracy to a considerable extent. Nevertheless, due to the inclusion of multi-head self-attention operations in FastViT, its inference speed remains marginally slower.

\subsection{Analysis of Model Energy and Power}
To ensure that the perception model can operate continuously for extended periods, particularly on edge devices, we also evaluate various models for their Energy consumed Per Sample (EPS) and AVerage Power (AVP) on both Orin and GTX 1650. 

As Table \ref{tab:energy_compare} presents, we once again conduct a rigorous comparison of our Achelous++ models capable of performing all five tasks simultaneously with other single-task models, single-modal models, and the fusion-based multi-task model, Efficient-VRNet-N. We observe that our model consumes a higher amount of EPS during inference, attributed to its three inputs, whereas other models typically have 1-2 inputs. Among them, MO-RDF-X-PN (S0) exhibits the lowest energy consumption per single sample on Orin, amounting to 741.1 joules. EV-GDF-X-PN (S0) shows a per-sample energy consumption of 940.2 joules on the GTX 1650. Remarkably, for the similarly radar-visual fusion-based model, Efficient-VRNet-N, these figures rise significantly to 1536.3 and 1925.7 joules, accompanied by a power consumption of 15.4 on Orin, surpassing our EF-GDF-X-PN (S0) and EV-GDF-X-PN (S0). It is worth noting that the inference speed of Efficient-VRNet-N is significantly lower than that of our model. Thus, Achelous++ successfully strikes a commendable balance between speed and power consumption.

For pure vision-based multi-task panoptic perception models, we observe that HybridNets has much higher EPS and AVP than our entire Achelous++ series. Similarly, YOLOP, another pure vision model capable of simultaneously performing three tasks, exhibits slightly lower power consumption than Achelous++ on Orin but higher on the GTX 1650. This further demonstrates Achelous++'s ability to achieve low-power inference while handling multi-modal inputs, fusion, and learning across five tasks.

Furthermore, our Achelous++ achieves power consumption levels that are comparable to, or even lower than, several single-task models. Additionally, we observe that models based on reparameterized structures, owing to their exceptionally fast inference speeds, exhibit relatively lower EPS values while achieving higher AVP.

\begin{table}
\setlength\tabcolsep{2.3pt}
\caption{Comparison of Energy Consumed Per Sample (EPS), Average Power (AVP) on WaterScenes Test Set.}
\centering
\label{tab:energy_compare}
\begin{tabular}{l|ccccc}  
\toprule   
  \textbf{Models} & \textbf{TN$\uparrow$} & \textbf{EPS$_\text{O}$(J)$\downarrow$} & \textbf{EPS$_\text{G}$(J)$\downarrow$}  & \textbf{AVP$_\text{O}$(W)$\downarrow$} & \textbf{AVP$_\text{G}$(W)$\downarrow$} \\
\midrule
\multicolumn{6}{c}{\texttt{Single-Task Models}} \\
\midrule
\multicolumn{6}{c}{Object Detection} \\
\midrule
  YOLOv8-Nano \cite{yolov8_ultralytics} &  1 & 243.0 & 250.9 & \textbf{12.6} & \textbf{42.1} \\
  YOLOv7-Tiny \cite{wang2023yolov7} &  1 & 170.5 & 262.0 & 13.8 & 63.2 \\
  YOLOX-Tiny \cite{ge2021yolox} &  1 & 238.6 & 255.4 & 12.9 & 50.9 \\
  YOLOv4-Tiny \cite{bochkovskiy2020yolov4} &  1 & \textbf{117.1} & \textbf{237.0} & 20.6 & 66.6 \\
  \midrule
  \multicolumn{6}{c}{Semantic Segmentation} \\
  \midrule
  Segformer-B0 \cite{xie2021segformer} &  1 & 428.9 & 720.22 & 19.9 & 64.3 \\
  PSPNet \cite{zhao2017pyramid} &  1 & \textbf{249.5} & \textbf{390.4} & \textbf{17.0} & \textbf{58.3} \\
  \midrule
  \multicolumn{6}{c}{Point Cloud Semantic Segmentation} \\
  \midrule
  PointNet & 1 & \textbf{83.9} & \textbf{100.5} & \textbf{13.4} & \textbf{52.4}\\
  PointNet++ & 1 & - & - & - & - \\
  PointNet-NN & 1 & 1205.8 & - & 35.4 & - \\
 \midrule
\multicolumn{6}{c}{\texttt{Multi-Task Models}} \\
\midrule
  \multicolumn{6}{c}{Panoptic Perception (Vision-based)} \\
  \midrule
  YOLOP \cite{wu2022yolop} &  3 & \textbf{582.6} & \textbf{787.0} & \textbf{14.0} & 64.7 \\
  HybridNets \cite{vu2022hybridnets} &  3 & 2922.6 & 3490.1 & 22.8 & \textbf{60.9} \\
\midrule
  \multicolumn{6}{c}{Panoptic Perception (Fusion-based)} \\
  \midrule
  Efficient-VRNet-N \cite{guan2023efficient} &  3 & 1536.3 & 1925.7 & 15.4 & \textbf{43.4} \\ 
\midrule
\textbf{\textit{S0}} \\
\midrule
  \textbf{EN-GDF-X-PN} &  \textbf{5} & 819.4 & 964.9 & 15.8 & \textbf{57.5}\\ 
  \textbf{EN-CDF-X-PN} &  \textbf{5} & 845.2 & 1120.3 & 16.1 & 58.5 \\
  \textbf{EF-GDF-X-PN} &  \textbf{5} & 876.7 & 1041.6 & \textbf{15.0} & 62.4 \\
  \textbf{EV-GDF-X-PN} &  \textbf{5} & 829.9 & \textbf{940.2} & \textbf{15.0} & 62.3 \\
  \textbf{MV-GDF-X-PN} &  \textbf{5} & 916.4 & 997.0 & 15.6 & 60.3 \\
  \textbf{MO-RDF-X-PN} &  \textbf{5} & \textbf{741.1} & 1144.8 & 20.4 & 63.4 \\
  \textbf{FV-RDF-X-PN} &  \textbf{5} & 810.7 & 1287.3 & 20.3 & 61.2 \\
\midrule
\textbf{\textit{S1}} \\
\midrule
  \textbf{EN-GDF-X-PN} & \textbf{5} & 895.6 & 1128.5 & 15.9 & \textbf{62.5} \\ 
  \textbf{EF-GDF-X-PN} & \textbf{5} & 1048.2 & 1348.4 & 16.1 & 63.3 \\
  \textbf{EV-GDF-X-PN} & \textbf{5} & 827.7 & \textbf{1089.2} & \textbf{15.4} & 62.7 \\
  \textbf{MV-GDF-X-PN} & \textbf{5} & 980.7 & 1209.8 & 16.7 & 62.6 \\
  \textbf{MO-RDF-X-PN} & \textbf{5} & \textbf{748.6} & 1167.5 & 20.2 & 64.1 \\
  \textbf{FV-RDF-X-PN} & \textbf{5} & 835.9 & 1361.2 & 19.9 & 62.9 \\
\midrule
\textbf{\textit{S2}} \\
\midrule
  \textbf{EN-GDF-X-PN} & \textbf{5} & 923.1 & 1283.2 & 16.4 & 64.5 \\
  \textbf{EF-GDF-X-PN} & \textbf{5} & 1293.8 & 1751.5 & 16.1 & 65.1 \\
  \textbf{EV-GDF-X-PN} & \textbf{5} & 1063.9 & 1405.4 & \textbf{15.8} & 64.2 \\
  \textbf{MV-GDF-X-PN} & \textbf{5} & 1034.7 & 1365.0 & 17.6 & 64.4 \\
  \textbf{MO-RDF-X-PN} & \textbf{5} &  \textbf{757.6} & \textbf{1210.1} & 20.4 & 64.2 \\
  \textbf{FV-RDF-X-PN} & \textbf{5} & 863.9 & 1457.5 & 20.3 & \textbf{62.8} \\

\bottomrule  
\end{tabular}\\
\end{table}

\subsection{Ablation Experiments}

\begin{table}
\setlength\tabcolsep{2.3pt}
\caption{Ablation Experiments of Achelous++ on the Basic Structures.}
\centering
\label{tab:ablation}
\begin{tabular}{l|lllll}  
\toprule   
  \textbf{Methods} & \textbf{mAP$_{\text{50-95}}$} & \textbf{mIoU$_\text{t}$} & \textbf{mIoU$_\text{d}$} & \textbf{mIoU$_\text{w}$} & \textbf{FPS$_{\text{O}}$}\\
  \midrule
  \textbf{MV-GDF-X-PN (S0) [baseline]} & \textbf{41.5} & \textbf{70.6} & \textbf{99.5} & 68.8 & 20.2 \\
  \midrule
  \textbf{-} SPPF & 40.8 & 69.8 & 99.1 & 67.7 & 20.8 \\
  \textbf{-} Dual Shuffle Attention & - & 68.7 & 98.9 & 65.2 & 20.4\\
  \textbf{-} Channel Shuffle (CS) & 41.2 & - & - & - & 20.2\\
  \textbf{-} Efficient Channel Attention & 40.6 & - & - & - & 20.2 \\
  \textbf{-} Radar Encoder & 40.4 & - & - & - & 21.3 \\
  Three segmentation branches & - & \textbf{70.9} & 99.5 & \textbf{69.0} & 18.2 \\
  One segmentation branch & - & 69.6 & 99.3 & 65.9 & \textbf{21.3} \\
\bottomrule  
\end{tabular}
\end{table}

Our ablation experiments include three aspects, basic structure of Achelous++ and radar encoder comparison.

Firstly, as Table \ref{tab:ablation} shows, in the absence of the SPPF module used for multi-scale feature fusion, we observe a decrease in several evaluation metrics, with a decline of 0.7 in the detection metric mAP$_{\text{50-95}}$ and a reduction in mIoU for various semantic segmentation tasks ranging from 0.4 to 1.1. Moreover, upon removal of the Dual Shuffle Attention module responsible for reweighing features for different segmentation tasks, a noticeable drop in mIoU is observed across all three segmentation tasks, ranging from 0.6 to 3.6. Furthermore, the exclusion of the ECA module has a detrimental impact on detection performance, highlighting the ECA's capacity to mitigate the adverse effects of radar clutter. Notably, the removal of the radar encoder leads to a substantial drop of 1.1 in mAP, underscoring the significant supplementary role of radar in enhancing visual detection. When the visual segmentation branch is divided into three, with each segmentation task employing a dedicated branch, we observe a slight improvement in mIoU for both object and waterline segmentation. However, the FPS experiences a significant decrease by two points. Conversely, when all three segmentation tasks share a single task head, we note a one-point increase in FPS. Nevertheless, the mIoU values for all three tasks show a decline, particularly in the case of waterline segmentation, which sees a reduction of nearly three points. 

Secondly, to validate the effectiveness of our Radar Convolution in comparison to conventional convolution, we conduct a comparison between MobileNetV2 and our Radar Encoder (RCNet) with the same number of channels as Table \ref{tab:rcnet} presents. We observe that MobileNetV2 exhibits a decrease in mAP$_{\text{50-95}}$ of 1.3 and a decrease in AR$_{\text{50}}$ of 2.1 compared to RCNet. Furthermore, its inference speed on Orin is 0.2 slower. These findings suggest that RCNet, with a similar parameter count, provides superior and faster modeling of point cloud features compared to conventional convolutions.

\begin{table}
\setlength\tabcolsep{3.5pt}
\caption{Ablation Experiments of Achelous++ on the Radar Encoder.}
\centering
\label{tab:rcnet}
\begin{tabular}{l|lll}  
\toprule   
  \textbf{Models} & \textbf{mAP$_{\text{50-95}}$} & \textbf{AR$_{\text{50}}$} & \textbf{FPS$_{\text{O}}$}  \\
  \midrule
  \textbf{MV-GDF-X-PN (RCNet)} & \textbf{41.5} & \textbf{45.6} & \textbf{20.0} \\
  \midrule
  MV-GDF-X-PN (MobileNetV2) & 40.2 & 43.5 & 19.8 \\
\bottomrule  
\end{tabular}
\end{table}

\begin{table}
\setlength\tabcolsep{2.5pt}
\caption{Inference Speed of Parallel Single-Task Models and Achelous++.}
\centering
\label{tab:speed_compare}
\begin{tabular}{l|c|ccc}  
\toprule   
  \textbf{Models} & \textbf{Tasks} & \textbf{EPS$_\text{O}$(J)$\downarrow$} & \textbf{AVP$_\text{O}$(W)$\downarrow$} & \textbf{FPS$_{\text{O}}\uparrow$}  \\

\midrule
  YOLOv4-Tiny \cite{ge2021yolox} & OD  & \multirow{4}[2]{*}{780.2}  & \multirow{4}[2]{*}{16.2} & \multirow{4}[2]{*}{20.5}  \\
  PSPNet-1 \cite{xie2021segformer} & SS \& DS &  \\
  PSPNet-2 \cite{zhao2017pyramid} & WS &   \\
  PointNet \cite{qi2017pointnet} & PC-SS &  \\
\midrule 
  \textbf{MO-RDF-X (S0)} & \textbf{OD \& SS \& DS}  &  \multirow{2}[2]{*}{\textbf{741.1}}  &  \multirow{2}[2]{*}{20.4} & \multirow{2}[2]{*}{\textbf{28.6}}  \\  
  &  \textbf{WS \& PC-SS} &      \\
\midrule 
  \textbf{EV-GDF-X (S0)} & \textbf{OD \& SS \& DS}  &  \multirow{2}[2]{*}{829.9}  &  \multirow{2}[2]{*}{\textbf{15.0}} & \multirow{2}[2]{*}{21.1}  \\  
  &  \textbf{WS \& PC-SS} &      \\
  
\bottomrule  
\end{tabular}
\end{table}

\subsection{Contrast Experiments}
The contrast experiments include two parts, the comparison of single-task model parallel systems with multi-task models and the comparison of multi-task training strategies. 

Firstly, as Table \ref{tab:speed_compare} presents, we execute the five different tasks concurrently on four of the fastest single-task models on Orin. We observe that the FPS of the parallel single-task models is 20.5, which is lower than the two different types of multi-task models in Achelous++ (reparameterized structure and hybrid backbone structure) with 28.6 and 21.1 FPS, respectively. The EPS of parallel single-task models is higher than that of MO-RDF-X-PN (S0) but lower than that of EV-GDF-X-PN (S0). Due to the extremely fast inference speed of MO-RDF-X-PN (S0), its AVP is higher than that of the parallel single-task models. However, the AVP of EV-GDF-X-PN (S0) is lower than that of the parallel single-task models. In summary, Achelous++ achieves faster speeds than multiple parallel single-task models at a lower power consumption.

Secondly, we conduct comparative experiments on multi-task optimization (Table \ref{tab:multi_task}). Initially, we employ MV-GDF-X-PN (S0), trained based on UW, as the baseline. We compare the performance of two different multi-task optimization strategies, MGDA and Aligned-MTL, on this model. While UW is based on balancing the weights of multiple losses, MGDA and Aligned-MTL rely on optimizing task-specific gradients. We observe that UW is the fastest in terms of training speed among the three strategies, with one epoch taking 9.4 minutes. However, it excels primarily in drivable-area segmentation. MGDA achieves the highest mIoU in semantic segmentation for object detection, but its performance lags behind in the other tasks compared to UW. Although Aligned-MTL outperforms the baseline in three of the tasks, with two of them reaching the highest scores, it  has the longest training time.

\begin{table}
\setlength\tabcolsep{3.5pt}
\caption{Comparison of Various Multi-Task Training Strategies.}
\centering
\label{tab:multi_task}
\begin{tabular}{l|ccccc}  
\toprule   
  \textbf{Strategies} & \textbf{mAP$_{\text{50-95}}$} & \textbf{mIoU$_{\text{t}}$} & \textbf{mIoU$_{\text{d}}$} & \textbf{mIoU$_{\text{w}}$} & \textbf{Mins$/$Epoch}$\downarrow$ \\
  \midrule
  \textbf{UW} \textbf{(baseline)} & 41.5 & 70.6 & \textbf{99.5} & 68.8 & \textbf{9.4} \\
  \midrule
  MGDA & 41.3 & \textbf{71.1} & 99.4 & 68.5 & 13.3\\
  Aligned-MTL & \textbf{41.6} & 70.9 & 99.3 & \textbf{69.4} & 17.1 \\
\bottomrule  
\end{tabular}
\end{table}

\subsection{Multi-Modal Structured Pruning}
As discussed in section \ref{sec:prune}, we construct the computation graph of Achelous++, group the nodes, and prune the input/output channels with respect to the ERK ratio. In Table \ref{tb:exp_prune}, we compare the pruned model with native Achelous++ with two different image backbones (MobileViT and EfficientFormer V2). MobileViT inherits the strong local feature representation of MobileNet and the global feature information of ViT, allowing it to outperform other backbones within the Achelous++ framework. Moreover, the model with EfficientFormer V2 is the largest among all models, whose hybrid structure is the most complex among all. Therefore, we choose to use the MobileViT backbone with an S2-size Achelous++ for the pruning process, and aim to achieve a better precision and speed trade-off than S0-size. In addition, EfficientFormer-based model is for validating the effectiveness of our pruning methods when confronting complex network structures.

\begin{table*}[ht]
  \centering
  \caption{Performance of Structural Pruned Achelous++ on Object Detection}
  \label{tb:exp_prune}
  \resizebox{\textwidth}{!}{%
\begin{tabular}{ll|rrrr|rrrr|rrrr|rr}
\toprule
\textbf{Models} & \textbf{Versions} & \multicolumn{2}{c}{\textbf{MACs(G)$\downarrow$}} & \multicolumn{2}{c|}{\textbf{Params(M)$\downarrow$}} & \multicolumn{2}{c}{\textbf{mAP$_\text{50-95}$$\uparrow$}} & \multicolumn{2}{c|}{\textbf{AR$_\text{50-95}$$\uparrow$}} & \multicolumn{2}{c}{\textbf{FPS$_{\text{O}}$$\uparrow$}} & \multicolumn{2}{c}{\textbf{FPS$_\text{{G}}$$\uparrow$}} & \multicolumn{2}{c}{\textbf{AVP$_\text{{O}}$(W)$\downarrow$}} \\ \hline\hline
\textbf{} & \textbf{S0} & 2.00 & -43.3\% & 2.5 & -60.3\% & 41.5 & -7.8\% & 45.6 & -6.6\% & 20.5 & 3.0\% & 61.5 & 28.2\% & - & - \\
\textbf{} & \textbf{S1} & 2.65 & -24.9\% & 3.7 & -41.0\% & 43.1 & -4.2\% & 47.2 & -3.3\% & 20.3 & 2.0\% & 52.2 & 9.7\% & - & -  \\
  \cmidrule(lr){2-16}
\textbf{MV-GDF-X} & \cellcolor[HTML]{EFEFEF}\textbf{S2 (baseline)} & \cellcolor[HTML]{EFEFEF}3.53 & \cellcolor[HTML]{EFEFEF}0.0\% & \cellcolor[HTML]{EFEFEF}6.2 & \cellcolor[HTML]{EFEFEF}0.0\% & \cellcolor[HTML]{EFEFEF}45.0 & \cellcolor[HTML]{EFEFEF}0.0\% &  \cellcolor[HTML]{EFEFEF}48.8 & \cellcolor[HTML]{EFEFEF}0.0\% & \cellcolor[HTML]{EFEFEF}19.9 & \cellcolor[HTML]{EFEFEF}0.0\% & \cellcolor[HTML]{EFEFEF}49.0 & \cellcolor[HTML]{EFEFEF}0.0\% & 17.5 & 0.0\% \\
\textbf{} & \textbf{ERK} & 1.98 & -44.1\% & \textbf{3.4} & \textbf{-45.1\%} & 43.0 & -4.4 \% & 46.9 & -3.9\% & 20.7 & 4.0\% & 59.9 & 22.2\% & 15.5 & -3.9\% \\
\textbf{} & \textbf{HA-SynFlow} & \textbf{1.97} & \textbf{-44.2\%} & \textbf{3.4} & -45.0\% & \textbf{43.4} & \textbf{-3.6\%}  & \textbf{48.0} & \textbf{-1.6\%} & \textbf{20.7} & \textbf{4.0\%} & \textbf{60.4} & \textbf{23.3\%} & \textbf{15.4} & \textbf{-4.2\%} \\
 \midrule
\textbf{} & \textbf{S0} & 2.16 & -46.6\% & 4.5 & -67.4\% & 37.4 & -7.7\% & 43.4 & -2.5\% & 20.8 & 30.8\% & 60.6 & 47.8\%  & - & - \\
\textbf{} & \textbf{S1} & 2.73 & -32.6\% & 7.1 & -48.4\% & 40.0 & -1.2\% & 43.8 & -1.6\% & 18.9 & 18.8\% & 53.3 & 30.0\%  & - & - \\
  \cmidrule(lr){2-16}
\textbf{EF-GDF-X} & \cellcolor[HTML]{EFEFEF}\textbf{S2 (baseline)} & \cellcolor[HTML]{EFEFEF}4.05 & \cellcolor[HTML]{EFEFEF}0.0\% & \cellcolor[HTML]{EFEFEF}13.7 & \cellcolor[HTML]{EFEFEF}0.0\% & \cellcolor[HTML]{EFEFEF}41.2 & \cellcolor[HTML]{EFEFEF}0.0\% & \cellcolor[HTML]{EFEFEF}44.5 & \cellcolor[HTML]{EFEFEF}0.0\% & \cellcolor[HTML]{EFEFEF}15.9 & \cellcolor[HTML]{EFEFEF}0.0\% & \cellcolor[HTML]{EFEFEF}41.0 & \cellcolor[HTML]{EFEFEF}0.0\% & 16.0 & 0.0\%\\
 & \textbf{ERK} & \textbf{2.12} & \textbf{-47.7\%} & \textbf{7.0} & \textbf{-48.9\%} & 39.0 & -3.7\% & \textbf{44.1} & \textbf{-0.9\%} & 20.8 & 31.0\% & 53.9 & 31.4\% & 14.1 & -11.8\% \\
 & \textbf{HA-SynFlow} & 2.12 & -47.5\% & 7.0 & -48.8\% & \textbf{39.3} & \textbf{-3.0\%} & 43.8 & -1.6\% & \textbf{21.2} & \textbf{33.3\%} & \textbf{54.3} & \textbf{32.4\%} & \textbf{14.0} & \textbf{-12.5\%} \\ \bottomrule
\end{tabular}%
}
\end{table*}

During the experiments, we constrain the pruned S2-sized models to have slightly lower MACs compared to the unpruned S0-sized models. This decision is made because, in the absence of hardware device-specific computational constraints, MACs are the primary metric for evaluating a model's speed. Furthermore, all pruning operations occur before neural network training. After initialization, structural pruning is carried out based on the calculated group sparsity. During pruning, we retain the parameters for two complete qkv (query, key, value) matrices of both backbones and only align the channels after pruning in adjacent modules.

The results in Table \ref{tb:exp_prune} demonstrate that MACs and Params do not have a one-to-one relationship. The size of module parameters and intermediate feature map in the computation graph collectively determine MACs. Therefore, reducing a network's size cannot solely rely on reducing the number of parameters; computational load should also be taken into account.

In the experiments with MV-GDF, all three performance metrics outperform the S0 and S1-sized models. For EF-GDF, the performance is slightly lower than S1 but better than S0. In the pruned networks, we significantly reduce the model size while minimizing accuracy loss. Surprisingly, the mAP$_{50}$ metric in MV-GDF suffers only a 1.6\% accuracy loss, and in EF-GDF, it even surpasses the original network. This aligns with the findings of another study on the impact of neural network pruning \cite{hooker2019compressed} where it is concluded that neural networks lose some fine-grained details after pruning, leading to greater accuracy drops in mAP$_{50-95}$, while mAP$_{50}$ remains relatively unaffected.

Moreover, we test the FPS achievable by the models. For resource-constrained devices like Jetson Orin, the pruned models exhibit better real-time performance than S0-sized models. In EF-GDF, the FPS is even 33.3\% higher than the original model. However, in the moderately resource-rich GTX1650 test, the pruned models only outperform S1 but still achieve 31.4\% and 32.4\% higher FPS compared to the original model. The primary reason for this difference lies in the fact that on the Orin device, the reduced model size allows for efficient GPU utilization, resulting in substantial computational improvements. Conversely, in the resource-rich GTX1650, certain small module computations are not fully utilized, leading to the described performance differential.

When we test the average power of pruned models on Orin, we can find that our pruning methods can remarkably reduce the power consumption with ratios from 3.9\% to 12.5\%, especially on EfficientFormer-based EF-GDF-X with the complex structure. It is noteworthy that for both logical model structures, our pruning strategies outperform ERK-based strategies, enabling pruned models to achieve lower AVP.

\begin{figure}[ht]
    \centering
    \includegraphics[width=1\linewidth]{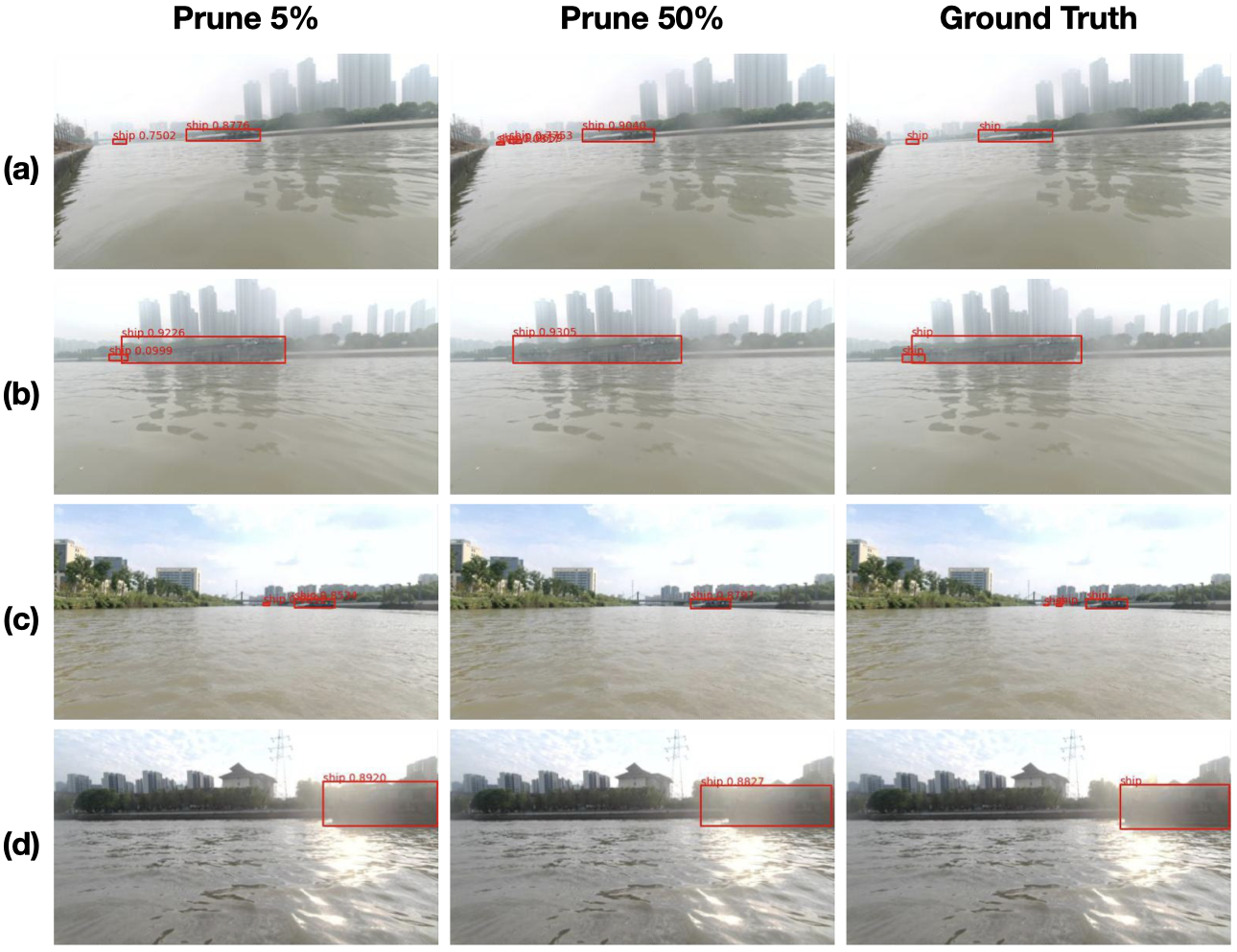}
    \caption{Pruning on radar encoder of MV-GDF-S2 (Left: prune less parameter of radar encoder; Middle: prune most parameter of radar encoder; Right: ground truth).}
    \label{fig:exp_prv}
\end{figure}

When we visualize models with \textbf{Pruned 5\% and 50\%} radar parameters, as shown in Fig. \ref{fig:exp_prv}, we observe a significant impact on the model's ability to detect challenging samples. In Fig. \ref{fig:exp_prv} (a), it becomes evident that a reduction in radar parameters leads to increased detection of low-confidence, small objects. This phenomenon arises because visual data alone may struggle to accurately identify very small pixel-sized targets, while radar can aid in confirming the presence of these targets. Similarly, in Fig. \ref{fig:exp_prv} (c), where smaller objects are in the visual field, \textbf{Pruned 50\%} tends to miss the object. Although \textbf{Pruned 5\%} misses one of the small objects, it captures another one with the assistance of the majority of radar parameters. As depicted in Fig. \ref{fig:exp_prv} (b), the visual system is more likely to overlook occluded objects in the absence of substantial radar modality assistance. Even in scenarios with intense sunlight, as seen in Fig. \ref{fig:exp_prv} (d), while the visual system can correctly detect the target, there are deviations in the bounding boxes of \textbf{Pruned 50\%}. 

Therefore, we can conclude that Achelous++ effectively balances the fusion of visual features and radar features. The network does not experience a decrease in detection performance due to excessive reliance on the weak modality. Instead, the weak modality effectively complements the detection of challenging samples for the strong modality.

\subsection{Visualization and Analysis of Prediction Results}
\begin{figure*}
    \centering
    \includegraphics[width=0.99\linewidth]{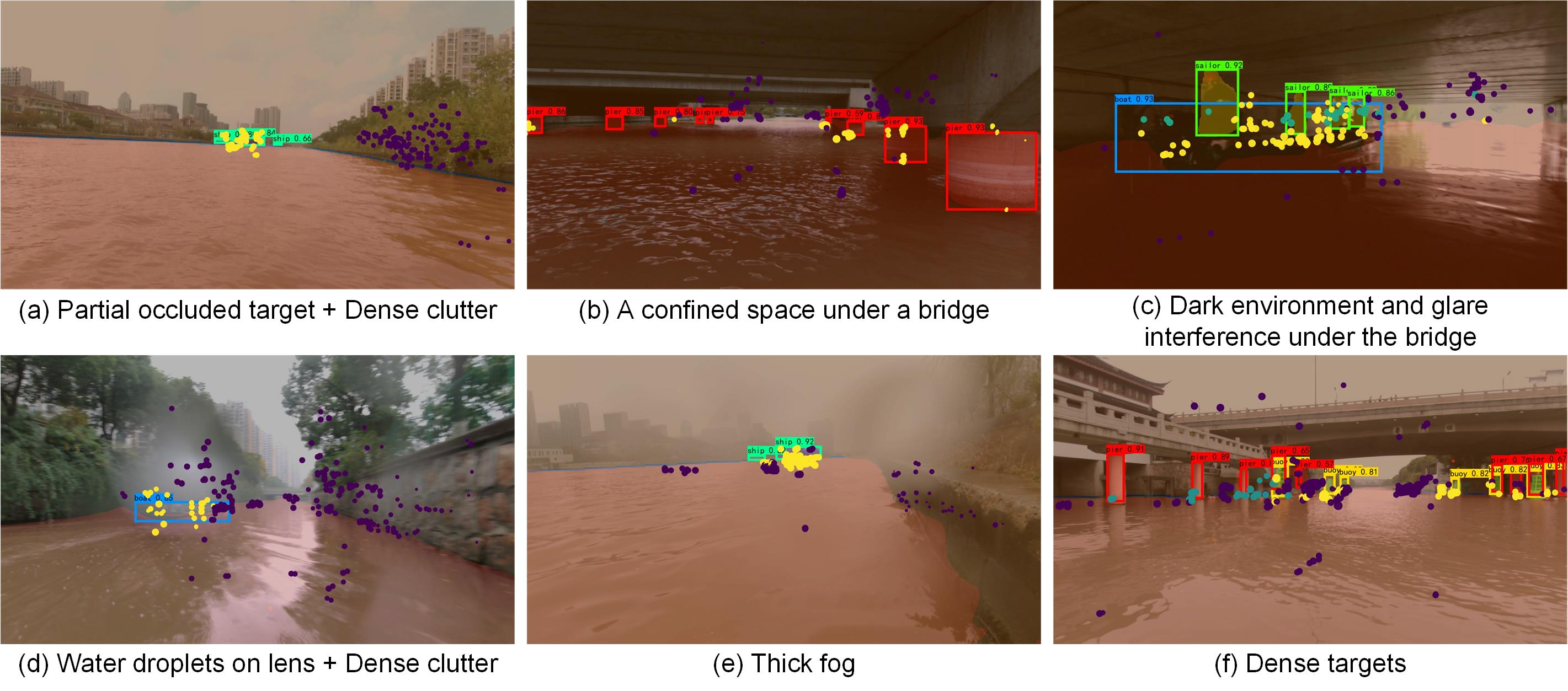}
    \caption{The prediction results of our MV-GDF-X-PN (S0) on the test set of WaterScenes, including several typical scenarios. Object detection predictions are in the format of bounding boxes. Objects of different categories are masked with masks of different colors. The drivable area is in red while the waterline prediction is in dark blue. Dark-blue point clouds denote the predicted clutter points while other colors refer to various targets.}
    \label{fig:achelous_output}
\end{figure*}

As Fig. \ref{fig:achelous_output} presents, we visualize the multi-task prediction results of MV-GDF-X-PN (S0) on the WaterScenes test set, which includes various complex scenarios such as (a) small and partially occluded targets with clutter along the riverbank, (b) enclosed and dark spaces under bridges, (c) distant glare interference, (d) lens droplets and water clutter on narrow river channels, (e) foggy conditions on the river, and (f) dense surface targets.

We observe that MV-GDF-X-PN (S0) excels in detecting partially occluded and small targets, demonstrating the effectiveness of the fusion method in making radar point cloud features compatible and complementary to visual features. Furthermore, in the presence of signal reflections in enclosed and narrow spaces, coupled with random interference from water clutter, our lightweight segmentation model still manages to accurately predict radar point cloud categories. In cases of complete visual failure, the supplementary radar features help prevent instances of missed detections.

\begin{figure*}
    \centering
    \includegraphics[width=0.99\linewidth]{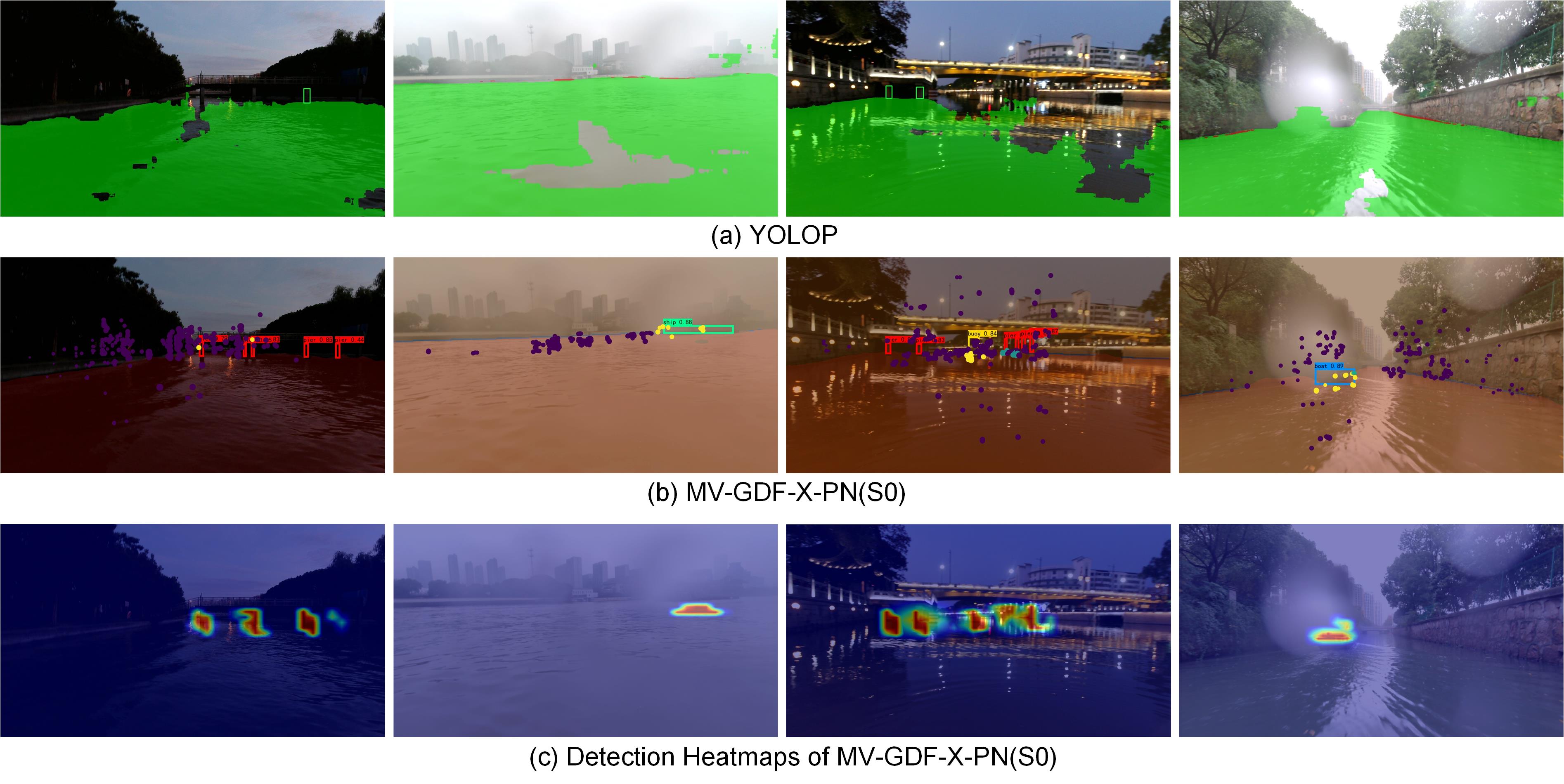}
    \caption{The comparison between YOLOP and MV-GDF-X-PN (S0). The corresponding heatmaps of detection heads are visualized.}
    \label{fig:compare_yolop}
\end{figure*}

Furthermore, as depicted in Fig. \ref{fig:compare_yolop}, we compare the predictions of MV-GDF-X-PN (S0) and the purely visual YOLOP in four extreme adverse scenarios. We visualize heatmaps of the object detection heads to observe whether the model makes correct detection decisions.

In the first column, within a very dark water environment, MV-GDF-X-PN (S0) detects multiple bridge piers ahead, while YOLOP only detects one. Further analysis of the radar point cloud, along with the heatmap in the last row, reveals that the heat regions were irregular, indicating that the radar point cloud provided prior knowledge of the presence of targets ahead and anchored the approximate location of the targets.

The second column shows a ship partially obscured by thick fog, rendering the ship invisible to the camera. YOLOP also fails to detect it, but MV-GDF-X-PN (S0) successfully detects the ship. The semantic segmentation of the radar point cloud is nearly perfect. Additionally, MV-GDF-X-PN (S0) shows higher quality in drivable area segmentation and waterline segmentation compared to YOLOP.

In the third column, the images display a nighttime scene with specular reflection interference on the water surface. YOLOP exhibits significant omission in drivable area segmentation, a problem considerably alleviated by MV-GDF-X-PN (S0). Moreover, false negatives in waterline segmentation and object detection are much higher for YOLOP compared to MV-GDF-X-PN (S0).

The final column demonstrates the presence of water droplets on the camera lens, obscuring parts of the image. YOLOP does not detect any objects under these conditions, but MV-GDF-X-PN (S0) identifies the presence of a moving boat obstructed by the water droplets. The point cloud segmentation results are also quite accurate. However, while the heatmap indicates that MV-GDF-X-PN (S0) recognizes the presence of two regions with potential targets, it fails to detect the sailor on the boat.

\section{Limitation Discussion and Future Works}
\label{sec:discussion}

\section{Conclusion}
\label{sec:conclusion}

The paper introduces Achelous++, an extensive water-surface panoptic perception framework, built upon the vision-radar multi-level fusion. It places a strong emphasis on developing energy-efficient, high-speed neural networks tailored for edge devices. Achelous++ is a versatile multi-task perception framework, accommodating a range of deep learning modules and featuring a convenient and effective multi-modal structural pruning library. We conduct a comprehensive performance analysis of our models, which demonstrates the successful optimization of precision, speed, and power consumption, ultimately leading to state-of-the-art results on the WaterScenes benchmark. We sincerely hope that Achelous++ can help promote the development of sustainable intelligent city with low-carbon emissions.


%

\section*{Acknowledgment}
The authors acknowledge XJTLU-JITRI Academy of Industrial Technology for giving valuable support to this article. This work is also partially supported by the XJTLU AI University Research Centre, and Jiangsu Province Engineering Research Centre of Data Science and Cognitive  Computation at XJTLU. In addition, it is partially funded by XJTLU-REF-21-01-002, Research Development Fund of XJTLU (RDF-19-02-23) and XJTLU Key Program Special Fund (KSF-A-17) as well as the Suzhou Municipal Key Laboratory for Intelligent Virtual Engineering (SZS2022004) and Suzhou Municipal Key Laboratory Broadband Wireless Access Technology (BWAT). This work receives financial support from Jiangsu Industrial Technology Research Institute (JITRI) and Wuxi National Hi-Tech District (WND).

\ifCLASSOPTIONcaptionsoff
  \newpage
\fi



%

\bibliographystyle{ieeetr}
\bibliography{bare_jrnl}



%

\vspace{-1.5cm}
\begin{IEEEbiography}[{\includegraphics[width=1in,height=1.25in,clip,keepaspectratio]{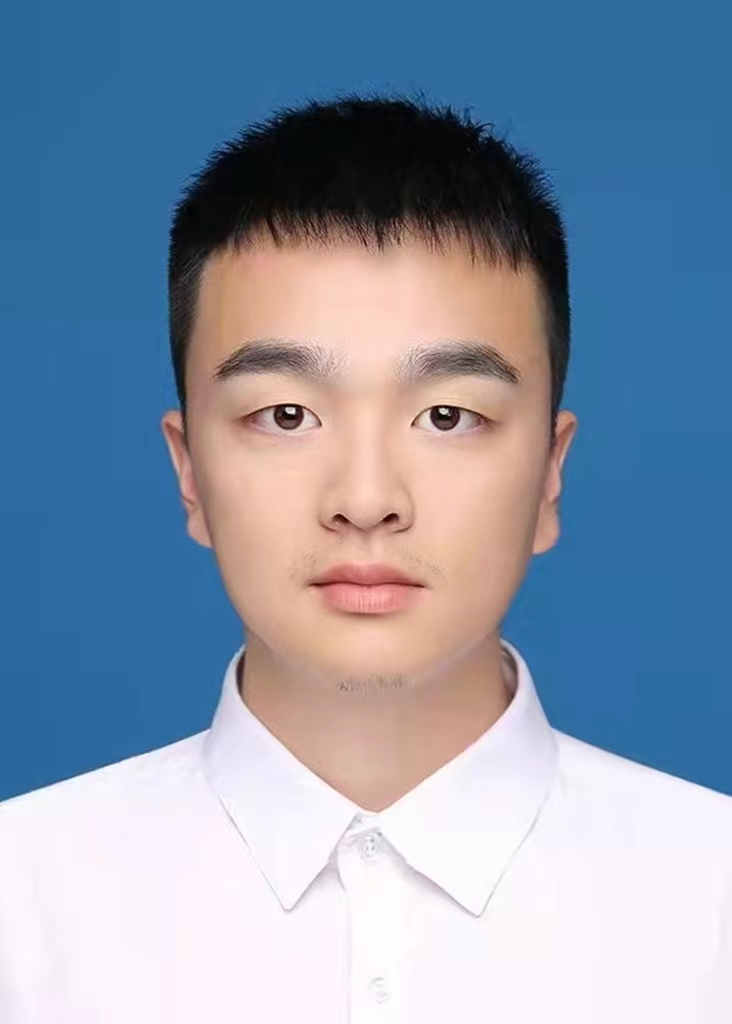}}]{Runwei Guan} (Student Member, IEEE) received his M.S. degree in Data Science from University of Southampton, Southampton, United Kingdom, in 2021. He is currently a joint Ph.D. student of University of Liverpool, Xi'an Jiaotong-Liverpool University and Institute of Deep Perception Technology, Jiangsu Industrial Technology Research Institute. His research interests include multi-modal perception, lightweight neural network, multi-task learning and statistical machine learning. He serves as a peer reviewer of IEEE TNNLS, IEEE TITS, ITSC and IEEE TCSVT, etc. He is also an active contributor on GitHub.
\end{IEEEbiography}
\vspace{-1.5cm}
\begin{IEEEbiography}[{\includegraphics[width=1in,height=1.25in,clip,keepaspectratio]{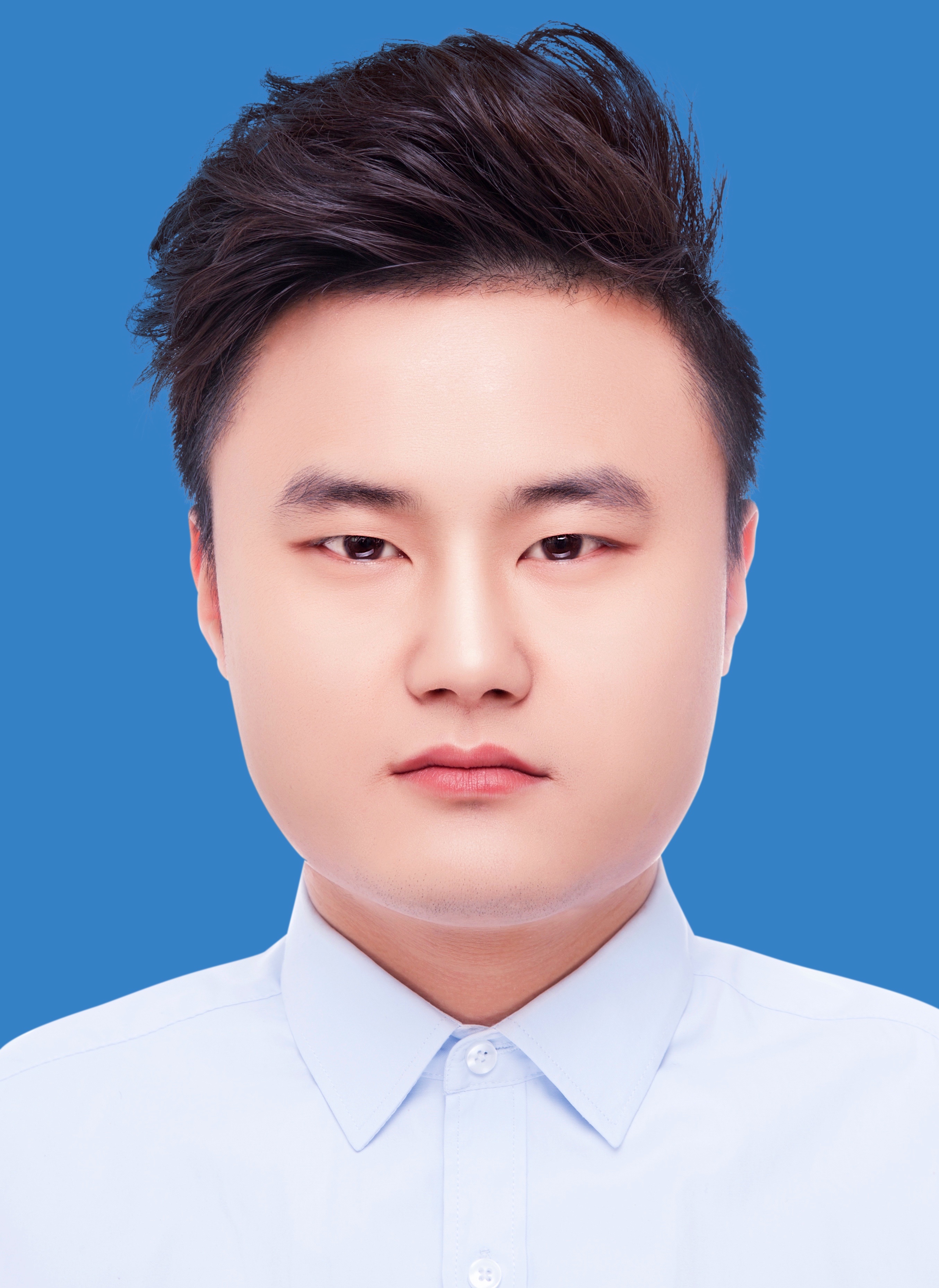}}]{Haocheng Zhao} received his B.E. and M.S. degree in 2019 and 2021, respectively, at Xi'an Jiaotong-Liverpool University, Jiangsu, China. He is currently a joint Ph.D. student of University of Liverpool, Xi'an Jiaotong-Liverpool University, Institute of Deep Perception Technology, and Jiangsu Industrial Technology Research Institute. His research interests include neural network pruning, radar perception, and robotics.
\end{IEEEbiography}
\vspace{-1.5cm}
\begin{IEEEbiography}[{\includegraphics[width=1in,height=1.25in,clip,keepaspectratio]{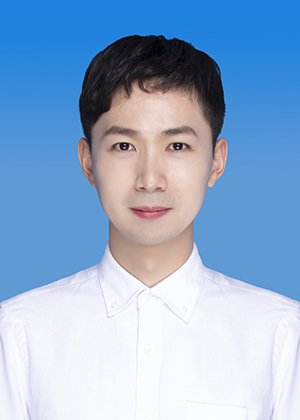}}]{Shanliang Yao} (Student Member, IEEE) received the B.E. degree in 2016 from the School of Computer Science and Technology, Soochow University, Suzhou, China, and the M.S. degree in 2021 from the Faculty of Science and Engineering, University of Liverpool, Liverpool, U.K. He is currently a joint Ph.D. student of University of Liverpool, Xi'an Jiaotong-Liverpool University and Institute of Deep Perception Technology, Jiangsu Industrial Technology Research Institute. His current research is centered on multi-modal perception using deep learning approach for autonomous driving. He serves as the peer reviewer of IEEE TITS, IEEE TIV, IEEE TCSVT, ICRA, etc.
\end{IEEEbiography}
\vspace{-1.5cm}
\begin{IEEEbiography}[{\includegraphics[width=1in,height=1.25in,clip,keepaspectratio]{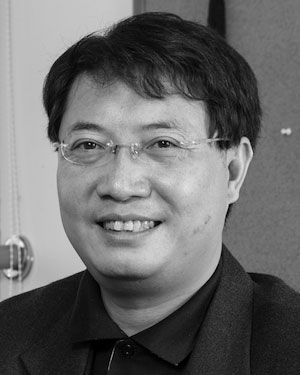}}]{Ka Lok Man}
(Member, IEEE), received the Dr.Eng. degree in electronic engineering from the Politecnico di Torino, Turin, Italy, in 1998, and the Ph.D. degree in computer science from Technische Universiteit Eindhoven, Eindhoven, The Netherlands, in 2006. He is currently a Professor in Computer Science and Software Engineering with Xi'an Jiaotong-Liverpool University, Suzhou, China. His research interests include formal methods and process algebras, embedded system design and testing, and photovoltaics.
\end{IEEEbiography}
\vspace{-1.5cm}
\begin{IEEEbiography}
[{\includegraphics[width=1in,height=1.25in,clip,keepaspectratio]{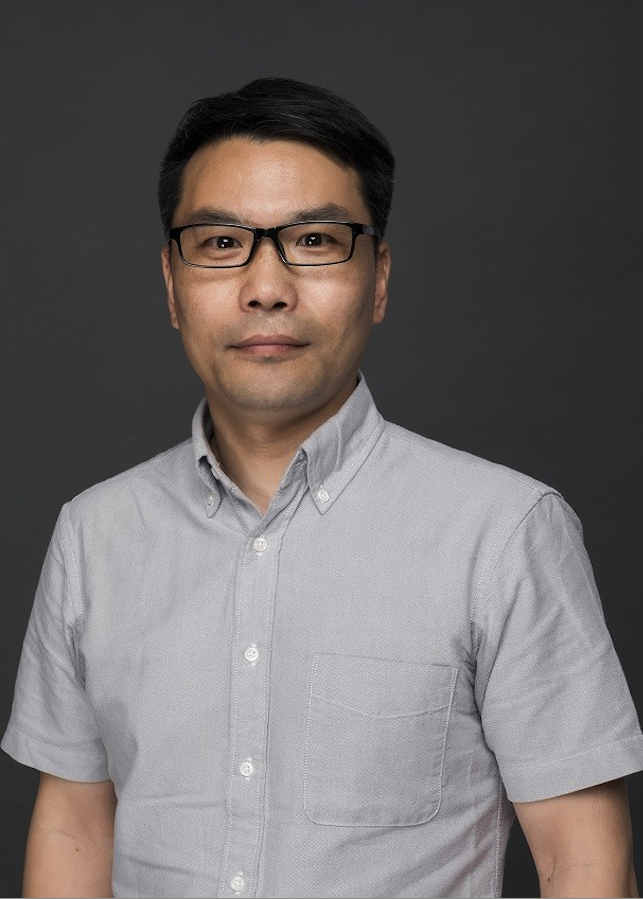}}]{Xiaohui Zhu}
(Member, IEEE) received his Ph.D. from the University of Liverpool, UK in 2019. He is currently an assistant professor, Ph.D. supervisor and Programme Director with the Department of Computing, School of Advanced Technology, Xi'an Jiaotong-Liverpool University. He focuses on advanced techniques related to autonomous driving, including sensor-fusion perception, fast path planning, autonomous navigation and multi-vehicle collaborative scheduling. 
\end{IEEEbiography}
\vspace{-1.5cm}
\begin{IEEEbiography}
[{\includegraphics[width=1in,height=1.25in,clip,keepaspectratio]{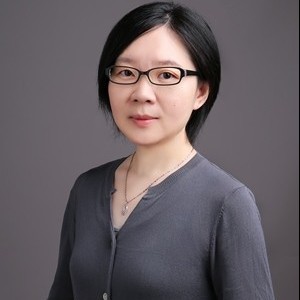}}]{Limin Yu}
(Member, IEEE) received the B.Eng. degree in telecommunications engineering and the M.Sc. degree in radio physics/underwater acoustic communications from Xiamen University, China, in 1999 and 2002, respectively, and the Ph.D. degree in telecommunications engineering from The University of Adelaide, Australia, in 2007. She worked with ZTE Telecommunications Company, Shenzhen, China, as a Software Engineer. She also worked with South Australia University and The University of Adelaide as a Research Fellow and a Research Associate. She has been with Xi’an Jiaotong-Liverpool University (XJTLU), since 2012; and is currently an Associate Professor. Her research interests include sonar detection, wavelet analysis, sensor networks, coordinated multi-AGV systems design, and medical image analysis.
\end{IEEEbiography}
\vspace{-1.5cm}
\begin{IEEEbiography}
[{\includegraphics[width=1in,height=1.25in,clip,keepaspectratio]{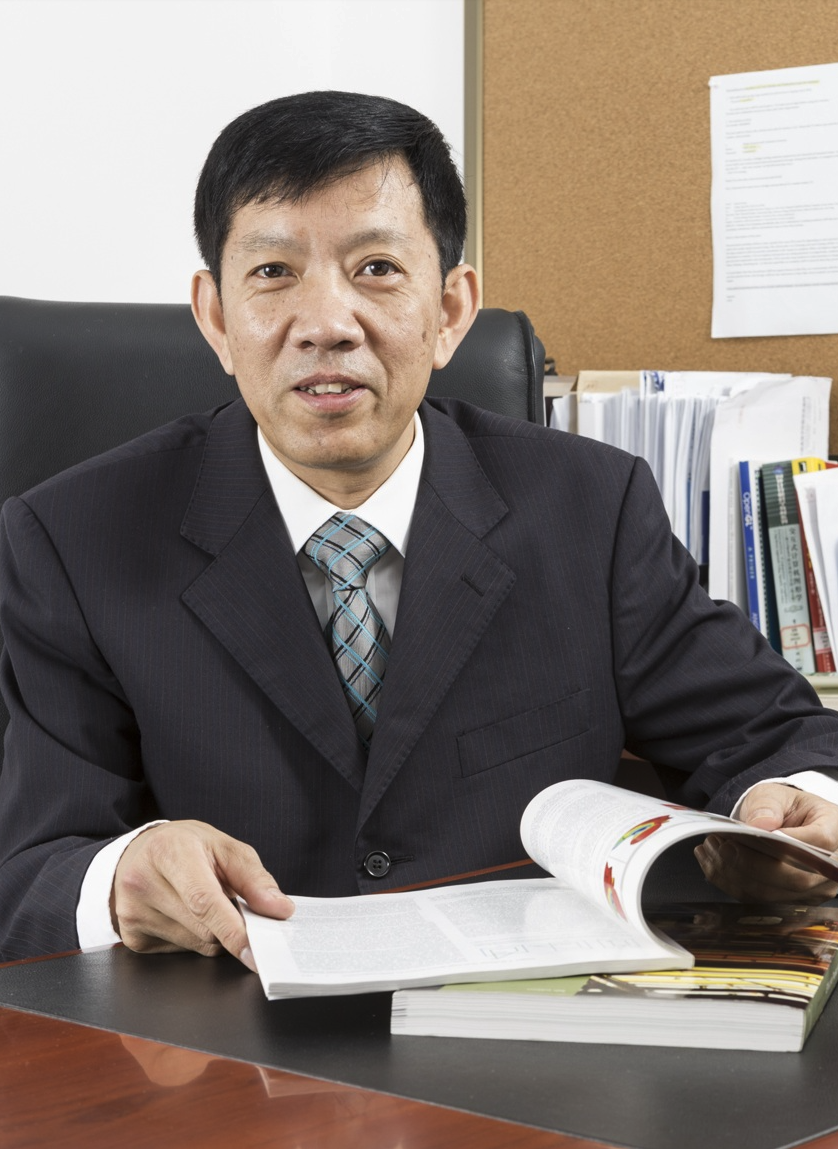}}]{Yong Yue}
Fellow of Institution of Engineering and Technology (FIET), received the B.Eng. degree in mechanical engineering from Northeastern University, Shenyang, China, in 1982, and the Ph.D. degree in computer aided design from Heriot-Watt University, Edinburgh, U.K., in 1994. He worked in the industry for eight years and followed experience in academia with the University of Nottingham, Cardiff University, and the University of Bedfordshire, U.K. He is currently a Professor and Director with the Virtual Engineering Centre, Xi'an Jiaotong-Liverpool University, Suzhou, China. His current research interests include computer graphics, virtual reality, and robot navigation.
\end{IEEEbiography}
\vspace{-2cm}
\begin{IEEEbiography}
[{\includegraphics[width=1in,height=1.25in,clip,keepaspectratio]{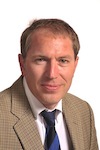}}]{Jeremy Smith}
(Member, IEEE), received the B.Eng. (Hons.) degree in engineering science and the Ph.D. degree in electrical engineering from the University of Liverpool, Liverpool, U.K., in 1984 and 1990, respectively.
Between 1984 and 1988, he conducted research on image processing and robotic systems with the Department of Electrical Engineering and Electronics, University of Liverpool, Liverpool, U.K., where he has been a Lecturer, Senior Lecturer, and Reader since 1988. He has held a Professorship position since 2006 in electrical engineering with the University of Liverpool. His research interests include automated welding, robotics, vision systems, adaptive control, and embedded computer systems.
\end{IEEEbiography}
\vspace{-2cm}
\begin{IEEEbiography}
[{\includegraphics[width=1in,height=1.25in,clip,keepaspectratio]{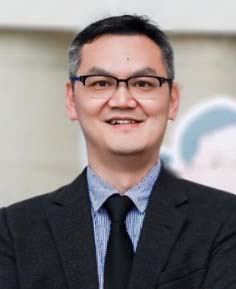}}]{Eng Gee Lim}
(Senior Member, IEEE) received the B.Eng. (Hons.) and Ph.D. degrees in Electrical and Electronic Engineering (EEE) from Northumbria University, Newcastle, U.K., in 1998 and 2002,
respectively. He worked for Andrew Ltd., Coventry, U.K., a leading communications systems company from 2002 to 2007. Since 2007, he has been with Xi'an Jiaotong–Liverpool University, Suzhou, China, where he was the Head of the EEE Department, and the University Dean of research and graduate studies. He is currently the School Dean of Advanced Technology, the Director of the AI University Research Centre, and a Professor with the Department of EEE. He has authored or coauthored over 100 refereed international journals and conference papers. His research interests are artificial intelligence (AI), robotics, AI+ health care, international standard (ISO/IEC) in robotics, antennas, RF/microwave engineering, EM measurements/simulations, energy harvesting, power/energy transfer, smart-grid communication, and wireless communication networks for smart and green cities. He is a Charted Engineer and a fellow of The Institution of Engineering and Technology (IET) and Engineers Australia. He is also a Senior Fellow of Higher Education Academy (HEA).
\end{IEEEbiography}
\vspace{-2cm}
\begin{IEEEbiography}
[{\includegraphics[width=1in,height=1.25in,clip,keepaspectratio]{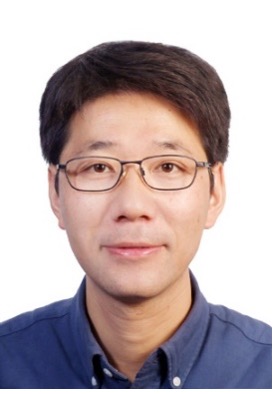}}]{Weiping Ding}
(Senior Member, IEEE) received the Ph.D. degree in computer science from the Nanjing University of Aeronautics and Astronautics, Nanjing, China, in 2013. He is currently a Full Professor with the School of Information Science and Technology, Nantong University, Nantong, China. In 2011, he was a Visiting Scholar with the University of Lethbridge, Lethbridge, AB, Canada. From 2014 to 2015, he was a Postdoctoral Researcher with Brain Research Center, National Chiao Tung University, Hsinchu, Taiwan. In 2016, he was a Visiting Scholar with the National University of Singapore, Singapore. From 2017 to 2018, he was a Visiting Professor with the University of Technology Sydney, Ultimo, NSW, Australia. He has authored or coauthored more than 150 scientific articles in refereed international journals, such as IEEE Transactions on Fuzzy Systems, IEEE Transactions on Neural Networks and Learning Systems, IEEE Transactions on Cybernetics, and IEEE Transactions on Evolutionary Computation. He has held 26 approved invention patents. He has coauthored two books. His authored/coauthored 15 papers have been selected as ESI Highly Cited Papers. His research interests include deep neural networks, multimodal machine learning, granular data mining, and medical images analysis.,Dr. Ding served/serves on the Editorial Board of Knowledge-Based Systems, Information Fusion, Engineering Applications of Artificial Intelligence and Applied Soft Computing. He is an Associate Editor for IEEE Transactions on Neural Networks and Learning Systems, IEEE Transactions on Fuzzy Systems, IEEE/CAA Journal of Automatica Sinica, IEEE Transactions on Intelligent Transportation Systems, IEEE Transactions on Emerging Topics in Computational Intelligence, IEEE Transactions on Artificial Intelligence, Information Sciences, Neurocomputing, Swarm and Evolutionary Computation, and so on.
\end{IEEEbiography}
\vspace{-2cm}
\begin{IEEEbiography}
[{\includegraphics[width=1in,height=1.25in,clip,keepaspectratio]{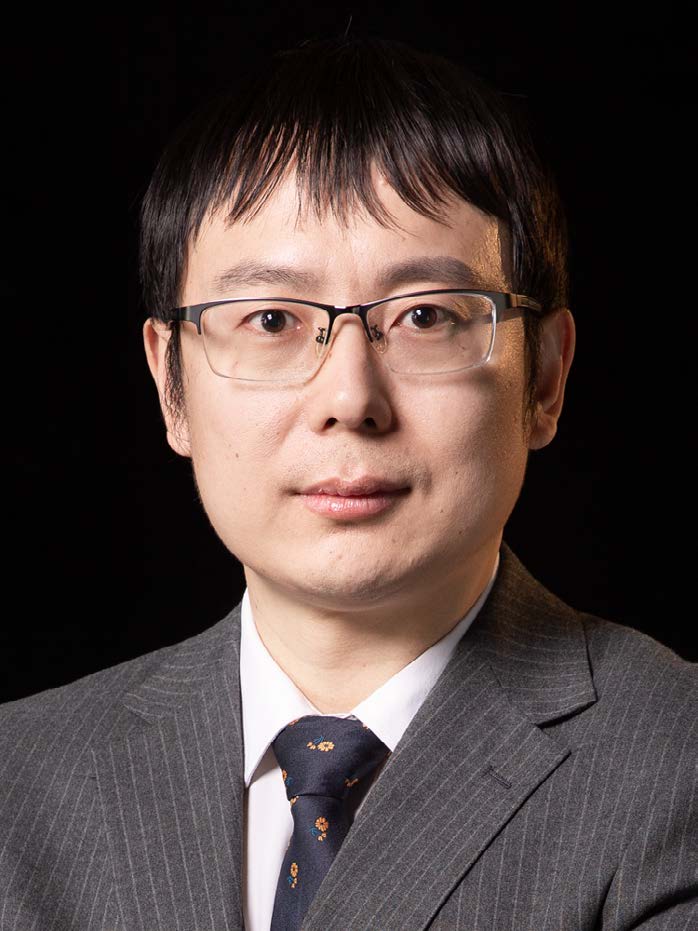}}]{Yutao Yue}
(Member, IEEE) was born in Qingzhou, Shandong, China, in 1982. He received the B.S. degree in applied physics from the University of Science and Technology of China, in 2004, and the M.S. and Ph.D. degrees in computational physics from Purdue University, USA, in 2006 and 2010, respectively. From 2011 to 2017, he worked as a Senior Scientist with the Shenzhen Kuang-Chi Institute of Advanced Technology and a Team Leader of the Guangdong ``Zhujiang Plan'' 3rd Introduced Innovation Scientific Research Team. From 2017 to 2018, he was a Research Associate Professor with the Southern University of Science and Technology, China. Since 2018, he has been the Founder and the Director of the Institute of Deep Perception Technology, JITRI, Jiangsu, China. Since 2020, he has been working as an Honorary Recognized Ph.D. Advisor of the University of Liverpool, U.K., and Xi'an Jiaotong-Liverpool University, China. He is the co-inventor of over 300 granted patents of China, USA, and Europe. He is also the author of over 20 journals and conference papers. His research interests include computational modeling, radar vision fusion, perception and cognition cooperation, artificial intelligence theory, and electromagnetic field modulation. Dr. Yue was a recipient of the Wu WenJun Artificial Intelligence Science and Technology Award in 2020.
\end{IEEEbiography}




\end{document}